\documentclass{article}

\PassOptionsToPackage{numbers, compress}{natbib}

 \usepackage[dblblindworkshop, final]{neurips_2026}
\workshoptitle{VLM4RWD}

\usepackage[utf8]{inputenc} %
\usepackage[T1]{fontenc}    %
\usepackage{url}            %
\usepackage{booktabs}       %
\usepackage{amsfonts}       %
\usepackage{nicefrac}       %
\usepackage{microtype}      %
\usepackage{xcolor}         %
\definecolor{linkblue}{RGB}{0,70,160}
\usepackage[colorlinks=true, linkcolor=linkblue, citecolor=linkblue, urlcolor=linkblue]{hyperref}

\usepackage[table]{xcolor}
\usepackage{pgfplots}
\pgfplotsset{compat=1.18} %
\usepgfplotslibrary{groupplots}
\usepackage[most]{tcolorbox}
\tcbset{promptbox/.style={
    colback=gray!5,
    colframe=gray!40,
    coltitle=black, %
    arc=1mm,
    boxrule=0.5pt,
    fonttitle=\bfseries\small,
    fontupper=\small\ttfamily,
    breakable,
    before upper={\setlength{\parskip}{0.6\baselineskip}\setlength{\parindent}{0pt}},
}}
\usepackage{booktabs}
\usepackage{tabularx}
\usepackage{multirow}
\usepackage{pdflscape}
\usepackage{pifont}
\usepackage{xspace}
\usepackage{makecell}
\usepackage{enumitem}
\usepackage[capitalize]{cleveref}
\usepackage{graphicx}
\usepackage{subcaption}
\usepackage{amsmath}        %
\usepackage{wrapfig}        %
\usepackage{titletoc}       %


\usepackage{pifont}
\newcommand{\cmark}{\ding{51}} %

\definecolor{bestgreen}{rgb}{0.69,0.91,0.62}
\definecolor{secondgreen}{RGB}{226,239,218}
\definecolor{pendingred}{rgb}{0.83,0.20,0.20}
\definecolor{lightgray}{rgb}{0.83,0.80,0.80}
\definecolor{lightviolet}{rgb}{0.90,0.86,0.93}
\definecolor{oursgreen}{HTML}{297708}
\definecolor{wrongred}{HTML}{A70A0A}
\definecolor{darkgray}{HTML}{404040}

\newcommand{\best}[1]{\boldmath\textbf{#1}}
\newcommand{\second}[1]{\underline{#1}}

\newcommand{\PPs}[1]{{\scriptsize$P{=}#1$}}   %
\newcommand{\qual}[1]{{\scriptsize (#1)}}      %
\newcommand{\bestoverall}[1]{\cellcolor{lightviolet}#1}

\newcommand{\PAR}[1]{\vskip 0pt \noindent{\bf #1~}}

\title{VLM-as-a-Judge for Person Re-Identification}

\author{%
  Bessie Dominguez-Dager$^{1}$\thanks{Corresponding author.} \quad
  Miguel Cazorla$^{1}$ \quad
  Vincent Lepetit$^{2}$ \\
  $^{1}$Institute for Computing Research, University of Alicante, Alicante, Spain \\
  $^{2}$LIGM, \'Ecole Nationale des Ponts et Chauss\'ees, IP Paris, Univ Gustave Eiffel, CNRS, France\\
  \texttt{\{bessie.dominguez, miguel.cazorla\}@ua.es}, \texttt{vincent.lepetit@enpc.fr}
}

\begin{document}

\maketitle

\begin{abstract}
  Cloth-changing person re-identification (CC-ReID) is still very challenging, and no method clearly dominates: methods that lead on one benchmark collapse on the others, because they rest on different identity cues. We show that a general-purpose vision-language model~(VLM), used zero-shot as a judge for reranking a short candidate list, beats the expert retrieval models it reranks by a wide and consistent margin. On the four scenarios of CHIRLA, a long-term CC-ReID benchmark that no publicly released Re-ID model is trained on, reranking the candidates of two frozen experts raises rank-1 on every scenario, by $12.9$ points on average over the best expert and $9.4$ over the best naive ensemble of the two, with no component fine-tuned on the target data. Merging both experts' shortlists matches or beats reranking the stronger expert's shortlist alone, at lower cost. Introducing a VLM does, however, come at a cost: almost all of the pipeline's FLOPs are spent on it. We therefore borrow a criterion from ensembling and invoke the VLM judge only where the two experts disagree at rank 1. This threshold-free gate removes $40\%$ of the calls, and a comparable share of total compute, with no significant drop in rank-1, trading $1.8$ to $3.8$ points of mAP. Judging remains the dominant cost, but on a benchmark where dedicated experts stall, we find a frozen general-purpose model to be the strongest judge of identity. The code is available at \url{https://github.com/bdager/JudgeReID}.
\end{abstract}

\section{Introduction}
\label{sec:intro}

Person re-identification (ReID) matches images of the same individual across camera views and locations~\cite{ye2021survey, LEE2025130763}.
Most methods address the short-term setting, where clothing and overall appearance remain unchanged (SC-ReID)~\cite{luo2019bot, he2021transreid, zhu2022pass, chen2023solider, clipreid, he2024instructreid, somers2024kpr}. However, practical deployments are long-term. Individuals reappear after weeks or months, with considerable changes in clothing, appearance, and environment.
This is ``cloth-changing ReID''~(CC-ReID), where clothing cues are unreliable by construction and methods must rely on identity-intrinsic evidence such as facial structure~\cite{arkushin2024geff, wu2024facecc, kim2025sapiensid}, body shape~\cite{seas, clip3dreid, chen2021learning3d, hong2021fsam}, or gait~\cite{jin2022gait, li23gait}. In this setting, accuracy remains far from the saturation reached in SC-ReID, and there is no consistent winner across benchmarks.
For example, under the cloth-changing~(CC) protocol, SapiensID~\cite{kim2025sapiensid} achieves state-of-the-art mAP on PRCC~\cite{yang2021prcc} (72.6\%), yet drops to 17.8\% on LTCC~\cite{qian2020ltcc}. In contrast, DIFFER~\cite{liang2025differ} reaches 31.6\% on LTCC but only 64.7\% on PRCC.
Such cross-benchmark reversals suggest that CC-ReID experts exploit different identity cues and thus make complementary errors. %

\begin{figure}[t]
  \centering
  \begin{subfigure}[b]{0.4\linewidth}
    \includegraphics[width=\linewidth]{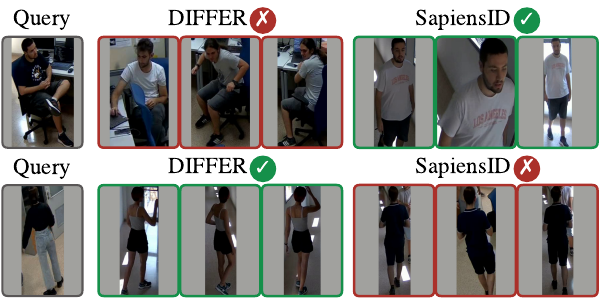}
  \end{subfigure}
  \hspace{0.02\linewidth}
  \begin{subfigure}[b]{0.25\linewidth}
    \includegraphics[width=\linewidth]{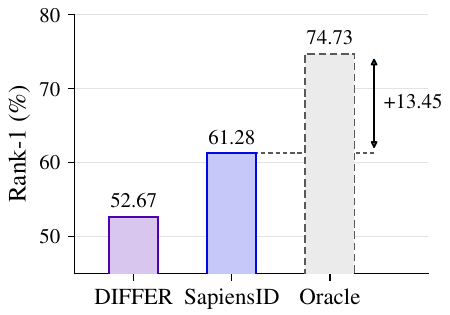}
  \end{subfigure}
  \begin{subfigure}[b]{0.3\linewidth}
    \includegraphics[width=\linewidth]{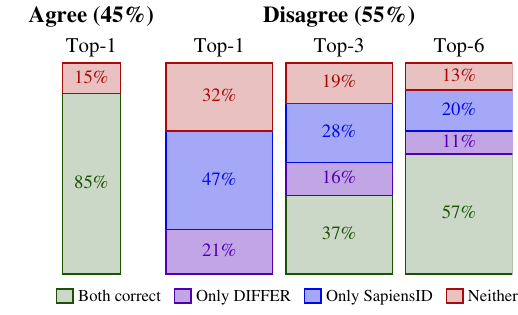}
  \end{subfigure}
  \caption{
\textbf{State-of-the-art CC-ReID experts make different mistakes}: We consider here
DIFFER and SapiensID on CHIRLA~\cite{dominguez2026chirla}, a challenging multi-camera scenario.
Left: two queries on which the experts disagree, with their top-3 candidates; \textcolor{oursgreen}{\textbf{green}} marks correct identities, \textcolor{wrongred}{\textbf{red}} incorrect ones. Each model succeeds where the other fails.
Middle: an oracle selecting the better model per query gains 13.5 rank-1 points over the best expert (dashed).
Right: queries split by top-1 agreement and by which model is correct. Agreement is a reliable correctness signal, and in the disagreement column only
13\% of queries have no correct candidate in either top-6 list.
In this paper, we introduce and evaluate different approaches to rerank the union of both shortlists with a VLM judge, on every query or, to save compute, only where the two disagree.
}
  \label{fig:oracle}
\end{figure}

This complementarity can be exploited at inference time, rather than committing to a single model.
We study this on SapiensID and DIFFER, the two experts mentioned above.
In~\cref{fig:oracle} we show that for this pair, an oracle selecting the better expert per query exceeds the stronger of the two by 13.5 rank-1 (R-1) points on the multi-camera scenario of CHIRLA~\cite{dominguez2026chirla}, a CC-ReID benchmark recorded over seven months.
In addition, we observe that agreement between their rank-1 predictions is a strong correctness signal, holding for 85\% of those agreements on this scenario. When they disagree, neither prediction is reliable, yet the true identity is often present in the union of their top-$k$ candidates.

We therefore build a training-free ensemble in which a vision-language model (VLM) arbitrates between two heterogeneous CC-ReID experts, and show that this frozen, general-purpose model is a better judge of identity than either expert it reranks.
Both models retrieve from the gallery independently.
Almost all of the pipeline's compute is spent on the judge, so we invoke it only where the two experts disagree at rank 1, which the agreement signal above justifies. Gating this way removes $40\%$ of the reranker calls and a comparable share of the total FLOPs, with no significant drop in R-1, trading $1.8$ to $3.8$ points of mAP.
Rather than arbitrating only between the two rank-1 predictions, the judge chooses from the union of both experts' top-$k$ candidates.

VLMs are generalists: their web-scale multimodal pretraining yields fine-grained visual understanding that transfers to biometric recognition~\cite{mllms_meet_reid}.
They are ill-suited as retrieval engines, since exhaustively scanning a large gallery is infeasible and their standalone accuracy trails task-specific models.
As judges over short candidate lists, however, VLMs make strong comparative decisions~\cite{chen2024mllm}, echoing the LLM-as-a-judge paradigm in the visual domain~\cite{zheng2023judging}.
Standard ensembling falls short here: score fusion~\cite{kittler1998combining, jain2005score} assumes that the two similarity scores are directly comparable, which independently trained embedding spaces do not provide, and rank aggregation such as reciprocal rank fusion (RRF)~\cite{rrf2009} averages the disagreement rather than resolving it.

The contributions of this work are as follows:
\begin{itemize}
  \item A zero-shot VLM judge that outperforms both experts and every naive ensemble of the two by a wide and consistent margin on all four CHIRLA scenarios, with no component trained or fine-tuned on the evaluated data.
  \item A threshold-free disagreement gate that skips the reranker on the queries where the experts agree, saving their compute with no significant drop in R-1 at the price of $1.8$ to $3.8$ points of mAP.
  \item An empirical study of how the candidate pool should be presented to the judge, covering pool size, pool ordering, and how the prompt binds each image to its role.
  \item Evidence on pool composition: merging both experts' shortlists matches or beats reranking a single expert's own shortlist at lower cost, even though deduplication often leaves the merged pool smaller.
\end{itemize}

\section{Related Work}
\label{sec:related}

\PAR{Clothing-Change Re-ID.}
CC-ReID methods remove clothing information from the identity representation at training time. One line uses language as a clothing-invariant supervisory signal, through cloth-agnostic CLIP alignment~\cite{clipcaf} or prompt learning over clothing and identity semantics~\cite{mmpl,mipl}. A related line factors identity from clothing in feature space, with DIFFER~\cite{liang2025differ} disentangling identity features from semantic cues over a frozen vision-language backbone and DRDnet~\cite{huang2026drdnet} modeling the relations between the resulting factors. Others synthesize clothing or pose variation at the feature~\cite{ccfa,pfaug,jfapl} or image level~\cite{ccaf,ccpa,nguyen24}, or ground identity in cues that survive a change of clothes, such as gait~\cite{li23gait}, implicit 3D shape~\cite{seas,clip3dreid} and skeleton dynamics~\cite{skeldyn}. Unsupervised variants pursue the same invariance without identity labels~\cite{cicl,japl,icsm} but remain far behind, with ICSM~\cite{icsm} reaching 13.5\% mAP on LTCC under the clothing-change protocol against 31.6\% for DIFFER~\cite{liang2025differ}.

SapiensID~\cite{kim2025sapiensid} takes a different route. Rather than modeling body structure explicitly, it unifies face and body recognition in one model, with identity supervision on WebBody, a web-scale corpus spanning diverse poses, scales and capture conditions. It transfers to CC-ReID off the shelf and, unlike the shape and gait methods above, exploits facial cues when visible. We build on DIFFER and SapiensID as our two experts.

\PAR{VLMs in person Re-ID.}
VLMs have already been used to some extent for ReID in two ways that differ in where the model intervenes. In the first one, language is a training-time signal only. The methods above, DIFFER~\cite{liang2025differ} among them, use text to shape a clothing-invariant embedding, as do pseudo-prompt optimization~\cite{clipreid}, caption-guided inversion~\cite{clipscgi}, language-guided feature enhancement~\cite{pedprompt,hcbn,idea} and diffusion-based augmentation~\cite{diffperson} outside the clothing-change setting. Retrieval then stays nearest-neighbor search over frozen features, and no language model sees a query at test time. VLM-ReID~\cite{lvlmreid} runs the model at test time, but only to emit an identity token used as the retrieval feature, so language still shapes the embedding rather than the ranking.

In the second way, the VLM is introduced in the retrieval loop. Text-to-image retrieval aligns free-form descriptions with gallery images~\cite{irra,unipt}, with VLMs supplying transferable captions~\cite{mllmt2i}, instruction-conditioned queries~\cite{he2024instructreid},
interactive refinement~\cite{hilreid} and dialogue-based retrieval~\cite{chatreid,convattr}.
Closest to us, Duan et al.~\cite{mllms_meet_reid} evaluate six off-the-shelf VLMs directly as ReID matchers under angle, illumination and clothing variation, and find that the prompt template moves the result. They score the VLM as a standalone matcher, while we invoke it as an arbiter over a short candidate list that two experts have already produced, and only where those experts disagree.

\PAR{Re-Ranking and model complementarity.}
Re-ranking refines a ranked list using gallery structure. $k$-reciprocal encoding~\cite{zhong17} remains the standard training-free baseline, revisited through adaptive graph-based neighbor search~\cite{chebgr}, while learned alternatives attend jointly over query and gallery~\cite{rrt21}. Pose2ID~\cite{pose2id} instead centralizes neighbor features as a training-free post-processing stage, validated only where clothing stays constant.
All of these refine the neighborhood structure of a single feature space, which is what clothing change destabilizes, and none recovers a match that the embedding misses entirely.
Re-ranking is therefore orthogonal to our question rather than an alternative to it: it could be applied on top of either expert or on the union of both, and we leave that combination to future work.
The methods we measure against are instead those that combine rankings from different models.

Combining heterogeneous rankers is much less explored in ReID. Score- and rank-level fusion such as RRF~\cite{rrf2009} treats models symmetrically and aggregates them on every query, whether or not they agree. Our correctness cross-tabulation shows that this spends effort where there is little to gain. On agreement queries the second model adds little, while disagreement queries carry nearly all of the oracle gap. Disagreement-gated arbitration turns that asymmetry into an invocation policy.

\section{Method}
\label{sec:method}

\subsection{Overview}
\label{sec:method:overview}

In our training-free arbitration ensemble, $M$ frozen ReID experts retrieve independently, and a frozen, general-purpose VLM judges the candidates they jointly propose. Throughout, we use $M = 2$.
We first select pretrained experts that present complementary error profiles. They succeed and fail under different conditions, such as whether the face is visible or whether pose changes between query and gallery.
All models are benchmarked off the shelf on datasets other than the ones they were trained on, so no component is fine-tuned on the target data.
Given a query $q$, each model $m \in \{1, \dots, M\}$ independently ranks the gallery $\mathcal{G} = \{g_1, \dots, g_N\}$ by similarity to $q$.
When all models predict the same identity for their top-ranked match, we return the ranking of a designated default model $m^\star$ unchanged.
When they disagree, we merge their top-$k$ matches into a candidate pool and ask the VLM to reorder it.
The final ranking places the $k_H$ top-ranked candidates from the reordered pool at the head, followed by the tail of the model that contributed the majority of that head.
We detail individual steps in the following subsections.

\subsection{Selective reranking}
\label{sec:method:reranking}

\PAR{Disagreement gating.}
Reranking every query with a VLM is expensive, and on queries the models already resolve it rarely changes the outcome while risking the demotion of a correct rank 1, as previously shown in \cref{fig:oracle}. We therefore invoke the VLM reranker only where the models disagree.
We assume an enrolled gallery whose identity labels are known at inference time, while the query identity is unknown. Let $\hat{y}_m(q)$ denote the enrolled identity label of the rank-1 gallery match returned by model $m$ for query $q$, and define the disagreement indicator
\begin{equation}
\delta(q) = \mathbf{1}\big[\, \big\lvert \{\hat{y}_m(q) : m = 1,\dots,M\} \big\rvert > 1 \,\big] \> ,
\label{eq:gate}
\end{equation}
which fires whenever the models do not all agree on the identity of their rank-1 match. Let $\mathcal{Q}_\delta = \{q \in \mathcal{Q} : \delta(q) = 1\}$ denote the gated subset of the query set $\mathcal{Q}$. The reranker is applied to $\mathcal{Q}_\delta$ and to no other query; for $q \notin \mathcal{Q}_\delta$ we return the ranking of the default model $m^\star$ unchanged.

\PAR{Candidate pool.}
Given a gated query, we pool the top-$k$ gallery items retrieved by each model:
\begin{equation}
\mathcal{C}(q) = \bigcup_{m=1}^{M} \left\{ \pi_m^{(r)}(q) \right\}_{r=1}^{k}, {k} \leq  |\mathcal{C}(q)| \leq {P} \> ,
\label{eq:pool}
\end{equation}
where $\pi_m(q)$ is model $m$'s ranking of the gallery, $\pi_m^{(r)}(q)$ its item at rank $r$ and $P = Mk$.
Each candidate carries the tag $\sigma(c) \subseteq \{1,\dots,M\}$ recording which models retrieved it; a candidate retrieved by several models carries all of them.
The pool is presented to the reranker under an ordering $\rho$, which we treat as a design choice rather than a fixed component. The reranker's sensitivity to $\rho$ is itself a property worth measuring, and we report it in~\cref{tab:pool_ordering}.
Merging rather than extending a single model's list to $M k$ items is what allows the pool to contain the correct match when one model ranks it highly and the other does not retrieve it at all.

\PAR{Prompt construction.}
We prompt the reranker to order $\mathcal{C}(q)$, from which the head of the final ranking is assembled.
Reranking requires the VLM to compare one query against several candidates in a single multi-image prompt.
In this setting, VLMs are susceptible to image position and recency effects~\cite{pos_bias25, paruchuri2026same}.
The VLM's answer is influenced by an image's position in the sequence and not only by its content. Thus, image identity becomes unreliable, with descriptions of one image applied to another.
We therefore build the prompt so that each image declares what it is, instead of leaving the model to infer it from where the image appears.
To do so, before any image is shown, the prompt gives all the instructions, including an \emph{anchor sentence} stating each image's role as query or candidate, and which images are not the query.
Then, each image carries a label written into the image itself rather than only in the surrounding text, following evidence that marks placed on an image are more reliably grounded than textual references to image order~\cite{som2023}.
The pool is presented in the order $\rho$ mentioned above.
\cref{sec:exp:prompt} evaluates the contribution of each prompt-construction choice.

\subsection{Final ranking}
\label{sec:method:final_rank}

The reranker orders the pool $\mathcal{C}(q)$ but leaves the remaining $N - |\mathcal{C}(q)|$ gallery images unranked, so a full ranking is assembled from the reranked head and one model's ranking.
We select that model per query rather than fixing it in advance.

\PAR{Tail selection.}
Of the ordering returned by the reranker we retain the first $k_H$ elements as the head $H(q) = (h_1, \dots, h_{k_H})$ and discard the rest.
Writing
\begin{equation}
c_m(q) = \left| \left\{\, i \leq k_H \;:\; m \in \sigma(h_i) \,\right\} \right|
\label{eq:contrib}
\end{equation}
for the number of head elements retrieved by model $m$, the majority contributor $\hat{m}(q) = \operatorname*{arg\,max}_{m} c_m(q)$ supplies the tail, with ties resolved in favor of $m^\star$.
The assembled ranking $R(q)$ is constructed from $\hat{\pi} = \pi_{\hat{m}(q)}(q)$ by replacing the first $k_H$ elements with $H(q)$ and resolving duplicates as described below. For ungated queries no pool exists and $R(q)$ is the full ranking of $m^\star$.

\PAR{Duplicate resolution.}
An image $h_i \in H(q)$ contributed by another model may sit at some position $p > k_H$ in $\hat{\pi}$, in which case it would appear twice in $R(q)$. Symmetrically, images that $\hat{\pi}$ originally placed in its own top-$k_H$ may have been displaced out of $H(q)$ by the reranker. Let $\mathcal{D}(q)$ be the sequence of displaced images $\{\hat{\pi}^{(r)}\}_{r=1}^{k_H} \setminus H(q)$ in order of original rank, and $\mathcal{V}(q)$ the vacated tail positions $\{p > k_H : \hat{\pi}^{(p)} \in H(q)\}$ in ascending order. Since $\hat{\pi}$ is a permutation of $\mathcal{G}$, every element of $H(q)$ occupies exactly one position in it, and $|\mathcal{D}(q)| = |\mathcal{V}(q)|$. Pairing them in order, the $j$-th vacated position receives the $j$-th displaced image,
\begin{equation}
R^{(r)}(q) =
\begin{cases}
h_r & \text{if } r \leq k_H, \\[2pt]
\mathcal{D}(q)_j & \text{if } r = \mathcal{V}(q)_j \> , \\[2pt]
\hat{\pi}^{(r)} & \text{otherwise,}
\end{cases}
\label{eq:assembly}
\end{equation}
so $R(q)$ contains every gallery image exactly once and has length $N$.

\section{Experiments}
\label{sec:experiments}

\PAR{Dataset and metrics.}
We evaluate our approach on CHIRLA~\cite{dominguez2026chirla}, a long-term Re-ID benchmark recorded over seven months in an indoor office with seven cameras and 22 identities, where participants changed clothing between sessions. We select it for several reasons. It is an evaluation-only benchmark, with no publicly released Re-ID model trained on it, so neither expert has seen its identities, cameras, or outfits. Its crops are high enough in resolution to preserve facial evidence.
It is far from saturated, so the disagreement regime our gate targets is well populated. Its four scenarios separate clothing change from viewpoint change.
Reappearance (Reapp.) and multi-camera (MC) stay within a session, while long-term (LT) and multi-camera long-term (MC-LT) span different days. We follow the benchmark protocol and compute metrics per subset before averaging. We adopt the closed-set setting and report R-1 and mAP.
When comparing two methods, we call a difference significant when an exact paired sign-flip test over identities gives $p < 0.05$, and give 95\% confidence intervals (CI) from a paired bootstrap over identities; \cref{sec:uncertainty} describes both, and \cref{tab:bootstrap_paired} reports them for every comparison in this section.
\cref{tab:datasets} gives the query and gallery sizes per scenario.

\PAR{Implementation details.}
All weights are frozen and nothing is trained or fine-tuned on CHIRLA.
The prompt configuration, $m^\star$ and $k_H$ were selected on the designated validation subset and then fixed for all evaluation subsets; $\rho$ was fixed arbitrarily, without consulting any performance number.
\cref{sec:exp:prompt} and \cref{sec:exp:ablations} therefore measure the effect of these choices on test.
The two experts are DIFFER~\cite{liang2025differ} and SapiensID~\cite{kim2025sapiensid}, using their released checkpoints, pre-trained on Celeb-ReID-light~\cite{celeb_reid} and WebBody12M~\cite{kim2025sapiensid} respectively.
We set $m^\star = \text{SapiensID}$, $\rho$ to the (DIFFER, SapiensID) block order, and $k_H = k$.
The judge is Qwen3.6-27B~\cite{qwen36_27b}, a dense VLM served with vLLM~\cite{vllm} behind an OpenAI-compatible API on two NVIDIA L40S GPUs, decoding greedily ($T = 0$) with reasoning mode disabled.
Outputs that are not a valid permutation of the pool are re-prompted at most twice, after which we fall back to $m^\star$.
\cref{supl:prompts} gives the prompt templates and \cref{fig:prompt_example} an example pool.
\cref{sec:exp:ablations} repeats the pipeline with Qwen3.6-35B-A3B~\cite{qwen36_35b_a3b}, a mixture-of-experts model.

\subsection{Main results}
\label{sec:exp:main}

\begin{wraptable}{r}{0.50\linewidth}
\centering
\setlength{\tabcolsep}{4pt}
\small
\vspace{-4.5mm}
\caption{\textbf{Gate analysis.}
\emph{Gated}: the experts' R-1 identities differ, so the judge is invoked.
\emph{Correct}: agreeing queries whose shared R-1 is right.
\emph{Oracle}: the better expert $m^\star$ against a per-query oracle.
\emph{Pool recall}: gated queries whose GT identity is in the merged $P{=}16$ pool;
the judge's R-1 ceiling.
}
\label{tab:headroom}
\vspace{1mm}
\resizebox{0.50\columnwidth}{!}
{\begin{tabular}{@{}lcccc@{}}
\toprule
          & Gated & Correct & Oracle (or.) & Pool recall \\
Scenario  & (\% all) & (\% agree) & R-1 $m^\star\!\to$ or. & (\% gated) \\
\midrule
LT      & 74.6 & 64.2 & $35.5 \to 50.1$ & 83.9 \\
MC-LT   & 65.2 & 70.9 & $43.4 \to 61.8$ & 94.0 \\
MC      & 54.9 & 85.5 & $61.3 \to 74.7$ & 90.2 \\
Reapp.  & 47.0 & 79.1 & $65.5 \to 75.5$ & 95.2 \\
\midrule
Avg.    & 60.4 & 74.9 & $51.4 \to 65.5$ & 90.8 \\
\bottomrule
\end{tabular}
}
\vspace{-5mm}
\end{wraptable}

\cref{tab:headroom} extends the oracle and error analysis of \cref{fig:oracle} to all four scenarios.
On average, the experts disagree at R-1 on $60.4\%$ of queries.
Where they agree, the shared prediction is correct $74.9\%$ of the time, above either
expert's overall R-1, so suppressing the judge there should cost little.
Agreement also fixes both experts to the same answer, so the entire oracle gap,
$51.4$ to $65.5$ R-1 on average, lies on the gated queries.
There, a merged $P = 16$ pool holds the correct identity $90.8\%$ of the time.
Hence the two choices the pipeline makes: invoke the judge only where the experts disagree, and
give it the merged union, since it holds the correct identity for the large majority of gated queries.

\begin{table}[t]
  \centering
  \caption{\textbf{Main results on CHIRLA}. Our pipeline against the two experts and standard ensembling. Per column, the best and second-best results, in \textbf{bold} and \underline{underlined}, respectively. $\Delta$ is the Avg.\ R-1 difference relative to SapiensID (ref.) with its 95\% confidence interval (CI) from a paired bootstrap over identities; $^{*}$ marks $p < 0.05$ under an exact sign-flip test (\cref{sec:uncertainty}).
  }
  \label{tab:main_headline}
  \resizebox{\linewidth}{!}{%
  \begin{tabular}{lccccccccccc}
    \toprule
    \multirow{2}{*}{Method} & \multicolumn{2}{c}{LT} & \multicolumn{2}{c}{MC-LT} & \multicolumn{2}{c}{MC} & \multicolumn{2}{c}{Reapp.} & \multicolumn{2}{c}{Avg.} & \multirow{2}{*}{\makecell{$\Delta$ R-1\\[-1pt]{\scriptsize [95\% CI]}}} \\
    \cmidrule(lr){2-3}\cmidrule(lr){4-5}\cmidrule(lr){6-7}\cmidrule(lr){8-9}\cmidrule(lr){10-11}
    & R-1 & mAP & R-1 & mAP & R-1 & mAP & R-1 & mAP & R-1 & mAP & \\
    \midrule
    DIFFER~\cite{liang2025differ}     & 30.89 & 31.20 & 42.56 & 48.51 & 52.67 & 53.79 & 54.15 & 55.45 & 45.07 & 47.24 & $-6.4$\,{\scriptsize$[-12.6, +0.2]$} \\
    SapiensID~\cite{kim2025sapiensid} & 35.50 & 35.01 & 43.44 & 46.63 & 61.28 & 57.94 & 65.50 & 61.22 & 51.43 & 50.20 & ref. \\
    \midrule
    Voting \qual{Borda}~\cite{aslam2001metasearch} & 38.63 & 36.17 & 45.81 & 50.27 & 60.33 & 59.27 & 62.84 & 61.00 & 51.90 & 51.68 & $+0.5$\,{\scriptsize$[-3.2, +4.3]$} \\
    RRF~\cite{rrf2009}  & 39.60 & 37.41 & 49.49 & 52.14 & 61.06 & 60.47 & 63.20 & 62.31 & 53.34 & 53.08 & $+1.9$\,{\scriptsize$[-1.8, +5.8]$} \\
    Score Fusion~\cite{jain2005score}                      & 40.95 & 38.31 & 48.60 & 51.78 & 64.47 & 61.75 & 65.93 & 63.90 & 54.99 & 53.94 & $+3.6$\,{\scriptsize$[+0.3, +7.0]$} \\
    \midrule
    Union + VLM \qual{gated} \PPs{6}      & \second{44.97} & \second{39.59} & \second{56.35} & \second{54.43} & \second{73.25} & \second{64.87} & \second{73.72} & \second{65.28} & \second{62.07} & \second{56.04} & $+10.6^{*}$\,{\scriptsize$[+7.0, +14.4]$} \\
    Union + VLM \qual{gated} \PPs{16}     & \bestoverall{\best{46.92}} & \bestoverall{\best{42.91}} & \bestoverall{\best{56.99}} & \bestoverall{\best{57.49}} & \bestoverall{\best{76.10}} & \bestoverall{\best{69.16}} & \bestoverall{\best{77.41}} & \bestoverall{\best{69.11}} & \bestoverall{\best{64.35}} & \bestoverall{\best{59.67}} & $+12.9^{*}$\,{\scriptsize$[+9.3, +16.8]$} \\
    \bottomrule
  \end{tabular}
}
\vspace{-2mm}
\end{table}

\PAR{Arbitration beats aggregation.}
The VLM here is not only refining one model's ranked list.
It decides which of two models' proposals to accept, acting as an arbitration ensemble method. In \cref{tab:main_headline}, we compare against standard ensembling over the experts' full rankings, so those baselines see the whole gallery where our judge sees $P$ candidates; \cref{sec:ensembles} gives the three formulations.
The best of them reaches $55.0$ average R-1, $3.6$ points above the better expert, a gain that is not significant.
The gated pipeline at $P = 16$ reaches $64.4$, $12.9$ points above SapiensID and $9.4$ above score fusion, both significant.
Even treating each person as the unit of analysis, the $12.9$-point gain has a 95\% CI of $[9.3, 16.8]$.
The margin holds scenario by scenario rather than only on the average, at least $11.4$ R-1 over the better expert and $6.0$ over the best ensemble on that scenario, with the same pattern on mAP, and it is significant in every scenario against both SapiensID and score fusion.
We attribute this to how the baselines combine the experts.
All three commit in advance to a fixed rule for combining the two lists, and score fusion must also combine scores that were never calibrated against each other, which min--max normalising does not make commensurable.
The judge instead looks at the crops and can decide differently on every query.

\subsection{What the union and the gate contribute}
\label{sec:exp:design}

\begin{figure}[t]
  \centering
  \includegraphics[width=0.9\linewidth, height=5cm]{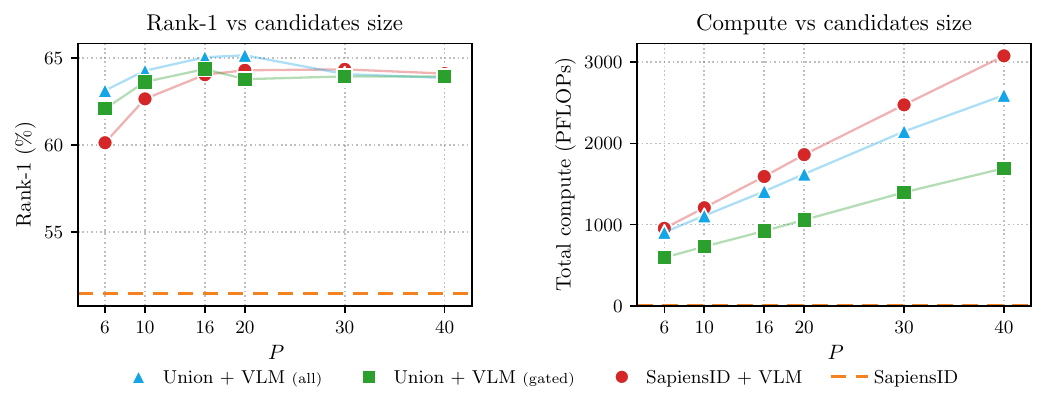}
  \vspace{-3mm}
  \caption{\textbf{Accuracy against compute as $P$ grows.} R-1 is averaged over the four scenarios and compute is the total FLOPs to run all four benchmarks once. All three pipelines sit above the SapiensID embedding-only baseline (dashed). The gated pipeline traces the same accuracy range at a fraction of the cost, since it calls the judge only where the experts disagree.
  }
  \label{fig:flops}
\vspace{-6mm}
\end{figure}

\PAR{The second expert pays for itself twice.}
Dropping it and reranking SapiensID's own top-$P$ is the natural alternative, plotted as \emph{SapiensID + VLM} in \cref{fig:flops}.
At matched $P$ and with the gate removed, the union costs less at every budget, $905$ PFLOPs against $954$, $1409$ against $1592$, and $2594$ against $3079$.
At $P=6$ it is ahead by $3.0$ R-1 and $3.9$ mAP; the mAP gain is significant and the R-1 gain borderline ($p = 0.054$).
At $P=16$ and $P=40$ the two are indistinguishable.
Pool size does not explain the gain, since deduplication leaves the merged pool \emph{smaller} than the budget, on average $5.1$ candidates against a full $6.0$ at $P=6$, so the judge does at least as well while seeing fewer crops, and that is also what makes it cheaper.
The advantage is only resolved at $P = 6$, likely because a single expert's longer shortlist already contains most of what the merge would add.
The second expert also supplies the disagreement signal, without which there is nothing to gate on, and gating is what brings the union down to $594$ PFLOPs at $P=6$.

\PAR{The gate trades mAP for compute.}
Against the same pipeline without it, gating has no significant effect on R-1 at any of the three budgets ($P = 6$: $-1.1$, $P = 16$: $-0.7$ and $P = 40$: $+0.1$).
In exchange the judge is skipped on the $40\%$ of queries where the experts agree, removing $35\%$ of the compute, which is the whole operating point, since the embedding stage is under $0.5\%$ of every pipeline's total.
Those two figures differ because the skipped queries are the cheaper ones: agreeing shortlists overlap more, so deduplication cuts deeper, leaving $4.5$ candidates against $5.5$ on the queries the judge does see.
The cost in mAP is larger, grows with the budget, and is significant at every budget: $1.8$, $3.3$ and $3.8$ points ($p \leq 2 \times 10^{-4}$, see \cref{tab:bootstrap_paired}).
On the queries the gate skips, the pipeline returns SapiensID's ranking unchanged, so nothing below rank 1 is reordered there, and the deeper the judge reorders elsewhere the more that untouched remainder costs.
A gate defined on rank-1 labels protects R-1 and leaves the rest of the list alone.

\PAR{Accuracy saturates before cost does.}
R-1 rises significantly from $P = 6$ to $P = 16$ and then flattens while cost keeps growing, and mAP keeps improving significantly to the largest budget tested.
Enlarging $P$ adds candidates below the current head, which lifts mAP mechanically but moves R-1 only when an added candidate displaces a wrong rank 1.
How the judge reorders the pool explains why R-1 stops rising.
At $P = 6$, restricted to queries whose ground truth is in the pool, the judge promotes it for $38.9$ to $62.3\%$ of queries and demotes it for $10.8$ to $22.2\%$, and within that subset R-1 rises from $26.1\%$ to $75.0\%$ on MC.
From $P = 16$ to $P = 40$ the demoted fraction grows in every scenario, by $1.9$ to $10.1$ points, while the promoted fraction falls by $0.7$ to $5.4$, so longer lists are reordered more aggressively without being reordered better.
Reliability degrades on the same axis, with corrected outputs growing from $0.2\%$ of queries at $P \leq 10$ to $32.0\%$ at $P = 40$ on LT under SapiensID + VLM (\cref{tab:format_failures}).

\subsection{Presenting the pool to the judge}
\label{sec:exp:prompt}

\begin{table}[t]
  \centering
  \begin{minipage}[t]{0.5\linewidth}
  \centering
  \caption{Prompt binding (Reapp. scenario). Images placed before or after the instructions, \emph{Anchor}: anchor sentence, \emph{Label}: role in the image pixels.
  }
  \label{tab:vlm_prompt_ablation}
  \small
  \resizebox{\columnwidth}{!}{
  \begin{tabular}{ccccrr}
    \toprule
    Img.\ before &Img.\ after & Anchor & Label & R-1 & mAP \\
    \midrule
    \cmark & &        &        & 65.98 & 60.59 \\
    \cmark &  & \cmark &        & 70.75 & 62.76 \\
    & \cmark  &        &        & 64.46 & 60.46 \\
    & \cmark  & \cmark &        & 72.99 & 63.73 \\
    & \cmark  & \cmark & \cmark & \textbf{73.72} & \textbf{65.28} \\
    \bottomrule
  \end{tabular}}
  \vspace{-4mm}
\end{minipage}
  \hfill
  \begin{minipage}[t]{0.46\linewidth}
  \centering
  \caption{Ablation of tail selection at $P=6$ on Union + VLM \qual{gated}. All values are mAP (\%) since R-1 is agnostic to the tail selection.}
  \label{tab:ablation}
\vspace{2.5mm}
  \small
  \setlength{\tabcolsep}{4pt}
  \begin{tabular}{lcccc}
    \toprule
    Tail & LT & MC-LT & MC & Reapp. \\
    \midrule
    DIFFER & 38.32 & 54.23 & 63.09 & 63.61 \\
    SapiensID & 38.51 & 52.95 & 63.05 & 64.39 \\
    Majority & \best{39.59} & \best{54.43} & \best{64.87} & \best{65.28} \\
    \bottomrule
  \end{tabular}
  \vspace{-4mm}
\end{minipage}
\end{table}

\PAR{Query binding dominates.}
\cref{tab:vlm_prompt_ablation} isolates the three choices that bind an image to its role, on Reappearance at $P = 6$.
All five variants carry the same task description, so the differences come from binding alone.
The anchor sentence adds $4.8$ to $8.5$ R-1 depending on image position; with the images before the instructions, all but one of the identities whose R-1 changed improved.
Burning the role into the image adds a further $1.6$ mAP, but its R-1 gain is not significant.
Image position has no significant effect, with or without the anchor sentence.
With the images after the instructions, the two interventions together add $9.3$ R-1 and $4.8$ mAP.
This matches what the position and recency bias literature predicts~\cite{pos_bias25, paruchuri2026same, som2023}, that the model needs to be told which of the $P + 1$ images is the query rather than to be told the task earlier.

\PAR{Task wording does not.}
Removing one semantic component at a time costs between $0.2$ and $1.7$ R-1 and under $0.9$ mAP, whether it is the attribute enumeration, the clothing prohibition and its escape hatch, or the step-by-step directive (\cref{tab:vlm_prompt_removals}), and none of these drops is significant.
Every one of those variants still beats every variant lacking the anchor sentence by at least $6.1$ R-1.
The gains therefore come from binding the query rather than from engineering the task description, which is the more transferable finding.

\begin{table}[t]
\centering
\setlength{\tabcolsep}{4.5pt}
\caption{Sensitivity to $\rho$ at $P=6$, with two orderings of $\mathcal{C}(q)$. $\Delta$ is the second order minus the first. Agreement is the fraction of \emph{gated} queries on which the two orders return the same identity sequence, at rank 1, over the retained head ($k_H=3$), and over the whole pool, whereas R-1 and mAP are over the full query set. In parentheses, chance agreement under random permutations of the same pool.
}
\label{tab:pool_ordering}
\resizebox{\linewidth}{!}{%
\begin{tabular}{lccccccccc}
\toprule
\multirow{2}{*}{Scenario} & \multicolumn{2}{c}{(DIFFER, SapiensID)} & \multicolumn{2}{c}{(SapiensID, DIFFER)} & \multicolumn{2}{c}{$\Delta$} & \multicolumn{3}{c}{Agreement \% (chance)} \\
\cmidrule(lr){2-3} \cmidrule(lr){4-5} \cmidrule(lr){6-7} \cmidrule(lr){8-10}
 & R-1 & mAP & R-1 & mAP & R-1 & mAP & @1 & @$k_H$ & @$P$ \\
\midrule
LT     & 44.97 & 39.59 & \best{45.94} & \best{39.88} & $+0.98$ & $+0.29$ & 69.6 \tiny(31.2) & 42.9 \tiny(4.9) & 31.2 \tiny(2.3) \\
MC-LT  & 56.35 & 54.43 & \best{58.50} & \best{55.11} & $+2.15$ & $+0.68$ & 69.2 \tiny(33.6) & 49.7 \tiny(6.1) & 39.7 \tiny(3.6) \\
MC     & \best{73.25} & 64.87 & 73.21 & \best{64.93} & $-0.04$ & $+0.06$ & 79.4 \tiny(33.2) & 51.5 \tiny(5.6) & 39.2 \tiny(3.0) \\
Reapp. & \best{73.72} & \best{65.28} & 73.43 & 65.01 & $-0.30$ & $-0.27$ & 82.1 \tiny(34.2) & 55.4 \tiny(5.9) & 41.1 \tiny(2.9) \\
\bottomrule
\end{tabular}%
}
\vspace{-6mm}
\end{table}

\PAR{Sensitivity to the pool ordering.}
\cref{tab:pool_ordering} varies $\rho$, keeping $\mathcal{C}(q)$ fixed and swapping the order of the per-expert top-$k$ blocks.
The two orders agree on the rank-1 identity for $69.2$ to $82.1\%$ of gated queries and on the full pool for $31.2$ to $41.1\%$, far above chance but with a large residual, consistent with concurrent work on order sensitivity in VLMs~\cite{paruchuri2026same}.
Aggregate accuracy is steadier, moving by $-0.3$ to $+2.1$ R-1 depending on the scenario, and the two orders are not significantly different in aggregate.
The order we use throughout was fixed in advance and is nominally the weaker of the two by $0.7$ R-1, so our reported results do not benefit from it.

\subsection{Ablations and analysis}
\label{sec:exp:ablations}
\label{sec:ablations}

\PAR{Tail rule.}
\cref{tab:ablation} compares fixing the tail to one expert against the per-query majority vote of \cref{eq:contrib}.
We report mAP only, since the tail rule applies only to gated queries, below the head the judge supplies, so it cannot move R-1, and ungated queries keep SapiensID's ranking under every rule.
The majority vote wins in every scenario, by a small margin that is significant on the average, $1.2$ and $1.3$ mAP over the DIFFER and SapiensID tails.
\cref{tab:ablation_ranks_sid} gives the breakdown over the ways of crediting a candidate that both models retrieved, which differ by at most $0.3$ mAP.

\begin{table}[t]
\centering
\caption{Results for two VLMs at $P=6$, gated pipeline.
\emph{Format} reports how often the response was not valid,
with \emph{Corr.} meaning repaired by the correction prompt, and \emph{Unrec.} still invalid after the
maximum of attempts, computed over the four Re-ID scenarios.
}
\label{tab:vlms_results}
\setlength{\tabcolsep}{4pt}
\resizebox{\linewidth}{!}{%
\begin{tabular}{lcccccccc|cc}
\toprule
\multirow{2}{*}{Method} & \multicolumn{2}{c}{LT} & \multicolumn{2}{c}{MC-LT} & \multicolumn{2}{c}{MC} & \multicolumn{2}{c}{Reapp.} & \multicolumn{2}{c}{Format (\%)} \\
\cmidrule(lr){2-3} \cmidrule(lr){4-5} \cmidrule(lr){6-7} \cmidrule(lr){8-9} \cmidrule(lr){10-11}
 & R-1 $\uparrow$ & mAP $\uparrow$ & R-1 $\uparrow$ & mAP $\uparrow$ & R-1 $\uparrow$ & mAP $\uparrow$ & R-1 $\uparrow$ & mAP $\uparrow$ & Corr. $\downarrow$ & Unrec. $\downarrow$ \\
\midrule
Qwen3.6-35B-A3B & 43.96 & 38.70 & 55.58 & 53.52 & 70.92 & 63.80 & 72.13 & 64.33 & 6.79 & 1.76 \\
Qwen3.6-27B & 44.97 & 39.59 & 56.35 & 54.43 & 73.25 & 64.87 & 73.72 & 65.28 & \best{0.23} & \best{0.12} \\
\bottomrule
\end{tabular}%
}
\end{table}

\PAR{A second VLM.}
\cref{tab:vlms_results} repeats the gated pipeline at $P = 6$ with Qwen3.6-35B-A3B.
It scores below Qwen3.6-27B in every scenario, significantly so on average ($1.4$ R-1 and $1.0$ mAP) but not in each scenario on its own, and shows the same pattern of gains.
It is also worse at following the output format, with higher correction and invalid-output rates.

\begin{figure}[t]
  \centering
  \includegraphics[width=9cm, height=4.5cm]{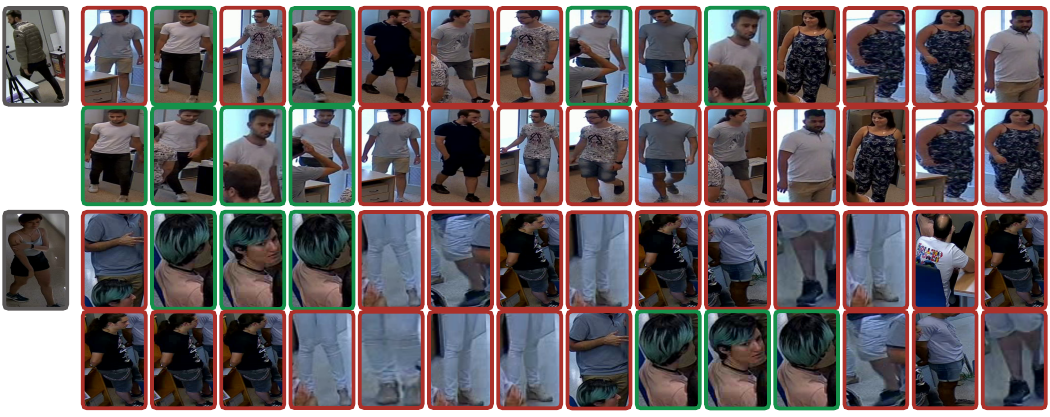}
  \caption{\textbf{Qualitative results} on two gated queries.
Query in \textcolor{darkgray}{\textbf{gray}}, \textcolor{oursgreen}{\textbf{green}} marks a correct identity and \textcolor{pendingred}{\textbf{red}} an incorrect one. For each query the upper row is the merged pool from DIFFER and SapiensID in the order shown to the judge, and the lower row is the order the judge returns. Above, the judge lifts four correct candidates to the head.
Below, it demotes the three correct candidates.
}
  \label{fig:qualitative}
\vspace{-4mm}
\end{figure}

\PAR{Qualitative results.}
\cref{fig:qualitative} shows two gated queries.
The first is a recovery, where the judge promotes an identity that neither expert ranked first.
The second is a failure, where the judge misjudges the gender of the candidates, treating long hair as an intrinsic female characteristic, and pushes the correct candidates down, from ranks 2 to 4 into positions 9 to 11.

\section{Conclusions}
\label{sec:conclusions}

We combine two frozen CC-ReID experts with complementary errors by having a general-purpose VLM rerank the union of their shortlists.
On CHIRLA, a benchmark recorded over seven months, the VLM judges identity better than either expert it reranks, at $64.4$ average R-1 and ahead of every naive ensemble of the two.
Merging the two shortlists costs less than reranking a single expert's list at the same budget and is not significantly worse at any budget, with a significant mAP gain at the smallest one.
The prompt matters as well, and its largest single gain comes from telling the judge which image is the query.
The judge dominates the pipeline's cost, so we invoke it only where the experts disagree; elsewhere their agreement is a reliable enough prediction.
That removes $40\%$ of the calls with no significant loss in R-1, at a cost of $1.8$ to $3.8$ mAP.
None of this requires training on the target data, calibrating one expert's scores against the other's, or a threshold to tune.

\PAR{Limitations.}
Our evaluation rests on CHIRLA's 22 identities.
How often the two experts disagree will therefore differ on other data, and with it how much the gate saves.
What the gate needs, though, is only that agreement predicts correctness.
Even gated, reranking stays the dominant cost, and the saving is paid for in mAP.
The judge is also order-sensitive, and its errors are not neutral: it sometimes groups candidates by coarse attributes such as apparent gender.
Nothing here is trained, however, so stronger experts or a stronger judge can be swapped in directly, on CHIRLA or on other CC-ReID benchmarks.

\begin{ack}
We thank Christian Henning for many insightful discussions, valuable ideas, and continuous support during the development of this work.

Bessie Dominguez-Dager is supported by the FPU predoctoral grant FPU23/02247 from the Spanish Ministry of Science, Innovation and Universities.
This work was carried out during a research stay at the Imagine group of the Laboratoire d'Informatique Gaspard-Monge (LIGM), École nationale des ponts et chaussées (France), funded by the FPU mobility grant FPUES23/00198 from the same ministry.
This work is also part of the project I+D+i PID2022-138453OB-I00, funded by MCIN/AEI/10.13039/501100011033/ and by ``ERDF, A way of making Europe''.

The authors declare no competing interests.
\end{ack}

{
    \small
    \bibliographystyle{ieeetr}
    \bibliography{references}

@inproceedings{arkushin2024geff,
  author={Arkushin, Daniel and Cohen, Bar and Peleg, Shmuel and Fried, Ohad},
  booktitle={2024 IEEE/CVF Winter Conference on Applications of Computer Vision Workshops (WACVW)},
  title={GEFF: Improving Any Clothes-Changing Person ReID Model Using Gallery Enrichment with Face Features},
  year={2024},
  volume={},
  number={},
  pages={143-153},
  doi={10.1109/WACVW60836.2024.00021}
}

@inproceedings{liang2025differ,
  author={Liang, Xin and Rawat, Yogesh S},
  booktitle={2025 IEEE/CVF Conference on Computer Vision and Pattern Recognition (CVPR)},
  title={DIFFER: Disentangling Identity Features via Semantic Cues for Clothes-Changing Person Re-ID},
  year={2025},
  volume={},
  number={},
  pages={13980-13989},
  doi={10.1109/CVPR52734.2025.01305}
}

@inproceedings{huang2026drdnet,
  title     = {Dual-stream Relation-modeling Disentanglement for Cloth-Changing Person Re-Identification},
  author    = {Huang, Shijuan and Ling, Hefei and Li, Zongyi and Li, Xu and Lv, Zhao},
  booktitle = {Proceedings of the AAAI Conference on Artificial Intelligence},
  year      = {2026},
  url       = {https://ojs.aaai.org/index.php/AAAI/article/view/37424}
}

@article{dominguez2026chirla,
  title   = {CHIRLA: Comprehensive High-resolution Identification and Re-identification for Large-scale Analysis},
  author  = {Dominguez-Dager, Bessie and Escalona, Felix and Gomez-Donoso, Francisco and Cazorla, Miguel},
  journal = {Scientific Data},
  volume  = {13},
  number  = {109},
  year    = {2026},
  url     = {https://www.nature.com/articles/s41597-025-06425-3},
  doi     = {10.1038/s41597-025-06425-3}
}

@inproceedings{kim2025sapiensid,
  author={Kim, Minchul and Ye, Dingqiang and Su, Yiyang and Liu, Feng and Liu, Xiaoming},
  booktitle={2025 IEEE/CVF Conference on Computer Vision and Pattern Recognition (CVPR)},
  title={SapiensID: Foundation for Human Recognition},
  year={2025},
  volume={},
  number={},
  pages={13937-13947},
  doi={10.1109/CVPR52734.2025.01301}
}

@inproceedings{zheng2023judging,
    author = {Zheng, Lianmin and Chiang, Wei-Lin and Sheng, Ying and Zhuang, Siyuan and Wu, Zhanghao and Zhuang, Yonghao and Lin, Zi and Li, Zhuohan and Li, Dacheng and Xing, Eric P. and Zhang, Hao and Gonzalez, Joseph E. and Stoica, Ion},
    title = {Judging LLM-as-a-judge with MT-bench and Chatbot Arena},
    year = {2023},
    publisher = {Curran Associates Inc.},
    address = {Red Hook, NY, USA},
    booktitle = {Proceedings of the 37th International Conference on Neural Information Processing Systems},
    articleno = {2020},
    numpages = {29},
    location = {New Orleans, LA, USA},
    series = {NIPS '23}
}

@inproceedings{chen2024mllm,
    author = {Chen, Dongping and Chen, Ruoxi and Zhang, Shilin and Wang, Yaochen and Liu, Yinuo and Zhou, Huichi and Zhang, Qihui and Wan, Yao and Zhou, Pan and Sun, Lichao},
    title = {MLLM-as-a-Judge: assessing multimodal LLM-as-a-Judge with vision-language benchmark},
    year = {2024},
    publisher = {JMLR.org},
    booktitle = {Proceedings of the 41st International Conference on Machine Learning},
    articleno = {254},
    numpages = {34},
    location = {Vienna, Austria},
    series = {ICML'24}
}

@article{ye2021survey,
  author={Ye, Mang and Shen, Jianbing and Lin, Gaojie and Xiang, Tao and Shao, Ling and Hoi, Steven C. H.},
  journal={IEEE Transactions on Pattern Analysis and Machine Intelligence},
  title={Deep Learning for Person Re-Identification: A Survey and Outlook},
  year={2022},
  volume={44},
  number={6},
  pages={2872-2893},
  doi={10.1109/TPAMI.2021.3054775}
}

@article{LEE2025130763,
    title = {Domain generalization for person re-identification: A survey towards domain-agnostic person matching},
    journal = {Neurocomputing},
    volume = {649},
    pages = {130763},
    year = {2025},
    issn = {0925-2312},
    doi = {https://doi.org/10.1016/j.neucom.2025.130763},
    url = {https://www.sciencedirect.com/science/article/pii/S0925231225014353},
    author = {Hyeonseo Lee and Juhyun Park and Jihyong Oh and Chanho Eom}
}

@inproceedings{clipreid,
    author = {Li, Siyuan and Sun, Li and Li, Qingli},
    title = {CLIP-ReID: exploiting vision-language model for image re-identification without concrete text labels},
    year = {2023},
    isbn = {978-1-57735-880-0},
    publisher = {AAAI Press},
    url = {https://doi.org/10.1609/aaai.v37i1.25225},
    doi = {10.1609/aaai.v37i1.25225},
    booktitle = {Proceedings of the Thirty-Seventh AAAI Conference on Artificial Intelligence and Thirty-Fifth Conference on Innovative Applications of Artificial Intelligence and Thirteenth Symposium on Educational Advances in Artificial Intelligence},
    articleno = {156},
    numpages = {9},
    series = {AAAI'23/IAAI'23/EAAI'23}
}

@inproceedings{he2021transreid,
  author={He, Shuting and Luo, Hao and Wang, Pichao and Wang, Fan and Li, Hao and Jiang, Wei},
  booktitle={2021 IEEE/CVF International Conference on Computer Vision (ICCV)},
  title={TransReID: Transformer-based Object Re-Identification},
  year={2021},
  volume={},
  number={},
  pages={14993-15002},
  doi={10.1109/ICCV48922.2021.01474}
}

@inproceedings{luo2019bot,
  author={Luo, Hao and Gu, Youzhi and Liao, Xingyu and Lai, Shenqi and Jiang, Wei},
  booktitle={2019 IEEE/CVF Conference on Computer Vision and Pattern Recognition Workshops (CVPRW)},
  title={Bag of Tricks and a Strong Baseline for Deep Person Re-Identification},
  year={2019},
  volume={},
  number={},
  pages={1487-1495},
  doi={10.1109/CVPRW.2019.00190}
}

@inproceedings{zhu2022pass,
    author="Zhu, Kuan
    and Guo, Haiyun
    and Yan, Tianyi
    and Zhu, Yousong
    and Wang, Jinqiao
    and Tang, Ming",
    editor="Avidan, Shai
    and Brostow, Gabriel
    and Ciss{\'e}, Moustapha
    and Farinella, Giovanni Maria
    and Hassner, Tal",
    title="PASS: Part-Aware Self-Supervised Pre-Training for Person Re-Identification",
    booktitle="Computer Vision -- ECCV 2022",
    year="2022",
    publisher="Springer Nature Switzerland",
    address="Cham",
    pages="198--214",
    isbn="978-3-031-19781-9"
}

@inproceedings{he2024instructreid,
  author={He, Weizhen and Deng, Yiheng and Tang, Shixiang and Chen, Qihao and Xie, Qingsong and Wang, Yizhou and Bai, Lei and Zhu, Feng and Zhao, Rui and Ouyang, Wanli and Qi, Donglian and Yan, Yunfeng},
  booktitle={2024 IEEE/CVF Conference on Computer Vision and Pattern Recognition (CVPR)},
  title={Instruct-ReID: A Multi-Purpose Person Re-Identification Task with Instructions},
  year={2024},
  volume={},
  number={},
  pages={17521-17531},
  doi={10.1109/CVPR52733.2024.01659}
}

@inproceedings{chen2023solider,
  author={Chen, Weihua and Xu, Xianzhe and Jia, Jian and Luo, Hao and Wang, Yaohua and Wang, Fan and Jin, Rong and Sun, Xiuyu},
  booktitle={2023 IEEE/CVF Conference on Computer Vision and Pattern Recognition (CVPR)},
  title={Beyond Appearance: A Semantic Controllable Self-Supervised Learning Framework for Human-Centric Visual Tasks},
  year={2023},
  volume={},
  number={},
  pages={15050-15061},
  doi={10.1109/CVPR52729.2023.01445}
}

@article{yang2021prcc,
    author={Yang, Qize and Wu, Ancong and Zheng, Wei-Shi},
    journal={ IEEE Transactions on Pattern Analysis \& Machine Intelligence },
    title={{ Person Re-Identification by Contour Sketch Under Moderate Clothing Change }},
    year={2021},
    volume={43},
    number={06},
    ISSN={1939-3539},
    pages={2029-2046},
    doi={10.1109/TPAMI.2019.2960509},
    url = {https://doi.ieeecomputersociety.org/10.1109/TPAMI.2019.2960509},
    publisher={IEEE Computer Society},
    address={Los Alamitos, CA, USA},
    month=jun
}

@inproceedings{pedprompt,
  author={Yang, Zexian and Wu, Dayan and Wu, Chenming and Lin, Zheng and Gu, Jingzi and Wang, Weiping},
  booktitle={2024 IEEE/CVF Conference on Computer Vision and Pattern Recognition (CVPR)},
  title={A Pedestrian is Worth One Prompt: Towards Language Guidance Person Re- Identification},
  year={2024},
  volume={},
  number={},
  pages={17343-17353},
  doi={10.1109/CVPR52733.2024.01642}
}

@inproceedings{lvlmreid,
  title={When large vision-language models meet person re-identification},
  author={Wang, Qizao and Li, Bin and Xue, Xiangyang},
  booktitle={ICASSP 2026-2026 IEEE International Conference on Acoustics, Speech and Signal Processing (ICASSP)},
  pages={10497--10501},
  year={2026},
  organization={IEEE}
}

@inproceedings{rrf2009,
    author = {Cormack, Gordon V. and Clarke, Charles L A and Buettcher, Stefan},
    title = {Reciprocal rank fusion outperforms condorcet and individual rank learning methods},
    year = {2009},
    isbn = {9781605584836},
    publisher = {Association for Computing Machinery},
    address = {New York, NY, USA},
    url = {https://doi.org/10.1145/1571941.1572114},
    doi = {10.1145/1571941.1572114},
    booktitle = {Proceedings of the 32nd International ACM SIGIR Conference on Research and Development in Information Retrieval},
    pages = {758–759},
    numpages = {2},
    location = {Boston, MA, USA},
    series = {SIGIR '09}
}

@inproceedings{rrt21,
  author={Tan, Fuwen and Yuan, Jiangbo and Ordonez, Vicente},
  booktitle={2021 IEEE/CVF International Conference on Computer Vision (ICCV)},
  title={Instance-level Image Retrieval using Reranking Transformers},
  year={2021},
  volume={},
  number={},
  pages={12085-12095},
  doi={10.1109/ICCV48922.2021.01189}
}

@inproceedings{qian2020ltcc,
    author="Qian, Xuelin
    and Wang, Wenxuan
    and Zhang, Li
    and Zhu, Fangrui
    and Fu, Yanwei
    and Xiang, Tao
    and Jiang, Yu-Gang
    and Xue, Xiangyang",
    editor="Ishikawa, Hiroshi
    and Liu, Cheng-Lin
    and Pajdla, Tomas
    and Shi, Jianbo",
    title="Long-Term Cloth-Changing Person Re-identification",
    booktitle="Computer Vision -- ACCV 2020",
    year="2021",
    publisher="Springer International Publishing",
    address="Cham",
    pages="71--88",
    isbn="978-3-030-69535-4"
}

@inproceedings{nguyen24,
  author={Nguyen, Vuong D. and Mantini, Pranav and Shah, Shishir K.},
  booktitle={2024 IEEE/CVF Conference on Computer Vision and Pattern Recognition Workshops (CVPRW)},
  title={Contrastive Clothing and Pose Generation for Cloth-Changing Person Re-Identification},
  year={2024},
  volume={},
  number={},
  pages={7541-7549},
  doi={10.1109/CVPRW63382.2024.00749}
}

@inproceedings{pose2id,
  author={Yuan, Chao and Zhang, Guiwei and Ma, Changxiao and Zhang, Tianyi and Niu, Guanglin},
  booktitle={2025 IEEE/CVF Conference on Computer Vision and Pattern Recognition (CVPR)},
  title={From Poses to Identity: Training-Free Person Re-Identification via Feature Centralization},
  year={2025},
  volume={},
  number={},
  pages={24409-24418},
  doi={10.1109/CVPR52734.2025.02273}
}

@inproceedings{clip3dreid,
  author={Liu, Feng and Kim, Minchul and Ren, Zhiyuan and Liu, Xiaoming},
  booktitle={2024 IEEE/CVF Conference on Computer Vision and Pattern Recognition (CVPR)},
  title={Distilling CLIP with Dual Guidance for Learning Discriminative Human Body Shape Representation},
  year={2024},
  volume={},
  number={},
  pages={256-266},
  doi={10.1109/CVPR52733.2024.00032}
}

@article{cicl,
    title = {Clothing-invariant contrastive learning for unsupervised person re-identification},
    journal = {Neural Networks},
    volume = {178},
    pages = {106477},
    year = {2024},
    issn = {0893-6080},
    doi = {https://doi.org/10.1016/j.neunet.2024.106477},
    url = {https://www.sciencedirect.com/science/article/pii/S0893608024004015},
    author = {Zhiqi Pang and Lingling Zhao and Chunyu Wang}
}

@inproceedings{zhong17,
  author={Zhong, Zhun and Zheng, Liang and Cao, Donglin and Li, Shaozi},
  booktitle={2017 IEEE Conference on Computer Vision and Pattern Recognition (CVPR)},
  title={Re-ranking Person Re-identification with k-Reciprocal Encoding},
  year={2017},
  volume={},
  number={},
  pages={3652-3661},
  doi={10.1109/CVPR.2017.389}
}

@inproceedings{hilreid,
  author={Qin, Yang and Chen, Chao and Fu, Zhihang and Peng, Dezhong and Peng, Xi and Hu, Peng},
  booktitle={2025 IEEE/CVF Conference on Computer Vision and Pattern Recognition (CVPR)},
  title={Human-centered Interactive Learning via MLLMs for Text-to-Image Person Re-identification},
  year={2025},
  volume={},
  number={},
  pages={14390-14399},
  doi={10.1109/CVPR52734.2025.01342}
}

@inproceedings{idea,
  author={Wang, Yuhao and Lv, Yongfeng and Zhang, Pingping and Lu, Huchuan},
  booktitle={2025 IEEE/CVF Conference on Computer Vision and Pattern Recognition (CVPR)},
  title={IDEA: Inverted Text with Cooperative Deformable Aggregation for Multi-modal Object Re-Identification},
  year={2025},
  volume={},
  number={},
  pages={29701-29710},
  doi={10.1109/CVPR52734.2025.02765}
}

@article{diffperson,
    author = {Niu, Ke and Yu, Haiyang and Qian, Xuelin and Fu, Teng and Li, Bin and Xue, Xiangyang},
    title = {Synthesizing efficient data with diffusion models for person re-identification pre-training},
    year = {2025},
    issue_date = {Feb 2025},
    publisher = {Kluwer Academic Publishers},
    address = {USA},
    volume = {114},
    number = {3},
    issn = {0885-6125},
    url = {https://doi.org/10.1007/s10994-024-06663-3},
    doi = {10.1007/s10994-024-06663-3},
    journal = {Mach. Learn.},
    month = feb,
    numpages = {25}
}

@inproceedings{ccaf,
    author = {Liu, Fangyi and Ye, Mang and Du, Bo},
    title = {Cloth-aware Augmentation for Cloth-generalized Person Re-identification},
    year = {2024},
    isbn = {9798400706868},
    publisher = {Association for Computing Machinery},
    address = {New York, NY, USA},
    url = {https://doi.org/10.1145/3664647.3680956},
    doi = {10.1145/3664647.3680956},
    booktitle = {Proceedings of the 32nd ACM International Conference on Multimedia},
    pages = {4053–4062},
    numpages = {10},
    location = {Melbourne VIC, Australia},
    series = {MM '24}
}

@inproceedings{chatreid,
  author={Bai, Yang and Ji, Yucheng and Cao, Min and Wang, Jinqiao and Ye, Mang},
  booktitle={2025 IEEE/CVF Conference on Computer Vision and Pattern Recognition (CVPR)},
  title={Chat-based Person Retrieval via Dialogue-Refined Cross-Modal Alignment},
  year={2025},
  volume={},
  number={},
  pages={3952-3962},
  doi={10.1109/CVPR52734.2025.00374}
}

@inproceedings{li23gait,
  author={Li, Weijia and Hou, Saihui and Zhang, Chunjie and Cao, Chunshui and Liu, Xu and Huang, Yongzhen and Zhao, Yao},
  booktitle={2023 IEEE/CVF Conference on Computer Vision and Pattern Recognition (CVPR)},
  title={An In-Depth Exploration of Person Re-Identification and Gait Recognition in Cloth-Changing Conditions},
  year={2023},
  volume={},
  number={},
  pages={13824-13833},
  doi={10.1109/CVPR52729.2023.01328}
}

@INPROCEEDINGS{mmpl,
author={Osamura, Kazuki and Uchida, Hidetsugu and Nie, Shijie and Abe, Narishige},
  booktitle={2024 IEEE International Joint Conference on Biometrics (IJCB)},
  title={Multi-Masked Prompt Learning For Clothing-Change Person Re-Identification},
  year={2024},
  volume={},
  number={},
  pages={1-8},
  doi={10.1109/IJCB62174.2024.10744448}
}

@article{mipl,
  author={Wei, Shengxun and Gao, Zan and Ma, Chunjie and Zhao, Yibo and Guan, Weili and Chen, Shengyong},
  journal={IEEE Transactions on Image Processing},
  title={Multiple Information Prompt Learning for Cloth-Changing Person Re-Identification},
  year={2025},
  volume={34},
  number={},
  pages={801-815},
  doi={10.1109/TIP.2025.3531217}
}

@article{japl,
  author={Pang, Zhiqi and Zhao, Lingling and Liu, Yang and Sharma, Gaurav and Wang, Chunyu},
  journal={IEEE Transactions on Information Forensics and Security},
  title={Joint Augmentation and Part Learning for Unsupervised Clothing Change Person Re-Identification},
  year={2025},
  volume={20},
  number={},
  pages={2944-2956},
  doi={10.1109/TIFS.2025.3550063}
}

@article{hcbn,
  author={Wang, Changshuo and Gao, Xingyu and Wu, Meiqing and Lam, Siew-Kei and He, Shuting and Tiwari, Prayag},
  journal={IEEE Transactions on Information Forensics and Security},
  title={Looking Clearer With Text: A Hierarchical Context Blending Network for Occluded Person Re-Identification},
  year={2025},
  volume={20},
  number={},
  pages={4296-4307},
  doi={10.1109/TIFS.2025.3558586}
}

@inproceedings{mllmt2i,
author={Tan, Wentao and Ding, Changxing and Jiang, Jiayu and Wang, Fei and Zhan, Yibing and Tao, Dapeng},
  booktitle={2024 IEEE/CVF Conference on Computer Vision and Pattern Recognition (CVPR)},
  title={Harnessing the Power of MLLMs for Transferable Text-to-Image Person ReID},
  year={2024},
  volume={},
  number={},
  pages={17127-17137},
  doi={10.1109/CVPR52733.2024.01621}
}

@inproceedings{unipt,
  author={Shao, Zhiyin and Zhang, Xinyu and Ding, Changxing and Wang, Jian and Wang, Jingdong},
  booktitle={2023 IEEE/CVF International Conference on Computer Vision (ICCV)},
  title={Unified Pre-training with Pseudo Texts for Text-To-Image Person Re-identification},
  year={2023},
  volume={},
  number={},
  pages={11140-11150},
  doi={10.1109/ICCV51070.2023.01026}
}

@inproceedings{pfaug,
    author = {Zhang, Jiajun and Li, Xin and Wu, Si and Xu, Yong and Wang, Yaowei},
    title = {Prior-Free Augmentation for Cloth-Changing Person Re-Identification},
    year = {2025},
    isbn = {9798400720352},
    publisher = {Association for Computing Machinery},
    address = {New York, NY, USA},
    url = {https://doi.org/10.1145/3746027.3755807},
    doi = {10.1145/3746027.3755807},
    booktitle = {Proceedings of the 33rd ACM International Conference on Multimedia},
    pages = {10612–10621},
    numpages = {10},
    location = {Dublin, Ireland},
    series = {MM '25}
}

@article{jfapl,
  title={Joint feature augmentation and posture label for cloth-changing person re-identification: L. Jiang et al.},
  author={Jiang, Liman and Zhang, Canlong and Wu, Lei and Li, Zhixin and Wang, Zhiwen and Wei, Chunrong},
  journal={Multimedia Systems},
  volume={31},
  number={2},
  pages={103},
  year={2025},
  publisher={Springer}
}

@inproceedings{seas,
  author={Zhu, Haidong and Budhwant, Pranav and Zheng, Zhaoheng and Nevatia, Ram},
  booktitle={2024 IEEE/CVF Conference on Computer Vision and Pattern Recognition (CVPR)},
  title={SEAS: ShapE-Aligned Supervision for Person Re-Identification},
  year={2024},
  volume={},
  number={},
  pages={164-174},
  doi={10.1109/CVPR52733.2024.00024}
}

@inproceedings{vllm,
    author = {Kwon, Woosuk and Li, Zhuohan and Zhuang, Siyuan and Sheng, Ying and Zheng, Lianmin and Yu, Cody Hao and Gonzalez, Joseph and Zhang, Hao and Stoica, Ion},
    title = {Efficient Memory Management for Large Language Model Serving with PagedAttention},
    year = {2023},
    isbn = {9798400702297},
    publisher = {Association for Computing Machinery},
    address = {New York, NY, USA},
    url = {https://doi.org/10.1145/3600006.3613165},
    doi = {10.1145/3600006.3613165},
    booktitle = {Proceedings of the 29th Symposium on Operating Systems Principles},
    pages = {611–626},
    numpages = {16},
    location = {Koblenz, Germany},
    series = {SOSP '23}
}

@inproceedings{chen2021learning3d,
  author={Chen, Jiaxing and Jiang, Xinyang and Wang, Fudong and Zhang, Jun and Zheng, Feng and Sun, Xing and Zheng, Wei-Shi},
  booktitle={2021 IEEE/CVF Conference on Computer Vision and Pattern Recognition (CVPR)},
  title={Learning 3D Shape Feature for Texture-insensitive Person Re-identification},
  year={2021},
  volume={},
  number={},
  pages={8142-8151},
  doi={10.1109/CVPR46437.2021.00805}
}

@inproceedings{hong2021fsam,
  author={Hong, Peixian and Wu, Tao and Wu, Ancong and Han, Xintong and Zheng, Wei-Shi},
  booktitle={2021 IEEE/CVF Conference on Computer Vision and Pattern Recognition (CVPR)},
  title={Fine-Grained Shape-Appearance Mutual Learning for Cloth-Changing Person Re-Identification},
  year={2021},
  volume={},
  number={},
  pages={10508-10517},
  doi={10.1109/CVPR46437.2021.01037}
}

@inproceedings{jin2022gait,
  author={Jin, Xin and He, Tianyu and Zheng, Kecheng and Yin, Zhiheng and Shen, Xu and Huang, Zhen and Feng, Ruoyu and Huang, Jianqiang and Chen, Zhibo and Hua, Xian-Sheng},
  booktitle={2022 IEEE/CVF Conference on Computer Vision and Pattern Recognition (CVPR)},
  title={Cloth-Changing Person Re-identification from A Single Image with Gait Prediction and Regularization},
  year={2022},
  volume={},
  number={},
  pages={14258-14267},
  doi={10.1109/CVPR52688.2022.01388}
}

@article{wu2024facecc,
  author = {Zhiyue Wu and Jianwei Ding and Qiong Zhang},
  title = {{FaceCC: facial feature combination network for cloth-changing person re-identification}},
  volume = {33},
  journal = {Journal of Electronic Imaging},
  number = {1},
  publisher = {SPIE},
  pages = {013045},
  year = {2024},
  doi = {10.1117/1.JEI.33.1.013045},
  URL = {https://doi.org/10.1117/1.JEI.33.1.013045}
}

@inproceedings{mllms_meet_reid,
  author = {Duan, Mengying and Li, He and Ye, Mang},
  title = {MLLMs Meet Person Re-identification},
  year = {2025},
  isbn = {9798400720352},
  publisher = {Association for Computing Machinery},
  address = {New York, NY, USA},
  url = {https://doi.org/10.1145/3746027.3758150},
  doi = {10.1145/3746027.3758150},
  booktitle = {Proceedings of the 33rd ACM International Conference on Multimedia},
  pages = {12247–12256},
  numpages = {10},
  location = {Dublin, Ireland},
  series = {MM '25}
}

@INPROCEEDINGS{celeb_reid,
  author={Huang, Yan and Wu, Qiang and Xu, Jingsong and Zhong, Yi},
  booktitle={2019 International Joint Conference on Neural Networks (IJCNN)},
  title={Celebrities-ReID: A Benchmark for Clothes Variation in Long-Term Person Re-Identification},
  year={2019},
  volume={},
  number={},
  pages={1-8},
  doi={10.1109/IJCNN.2019.8851957}
}

@INPROCEEDINGS{pos_bias25,
  author={Tian, Xinyu and Zou, Shu and Yang, Zhaoyuan and Zhang, Jing},
  booktitle={2025 IEEE/CVF Conference on Computer Vision and Pattern Recognition (CVPR)},
  title={Identifying and Mitigating Position Bias of Multi-image Vision-Language Models},
  year={2025},
  volume={},
  number={},
  pages={10599-10609},
  doi={10.1109/CVPR52734.2025.00991}
}

@inproceedings{convattr,
    author    = {Zou, Hao and Zhang, Runqing and Ding, Jin and Zhou, Xue and Zou, Jianxiao and Cai, Mingzhu},
    title     = {Tackling Alignment Ambiguity in Person Retrieval through Conversational Attribute Mining},
    booktitle = {Proceedings of the IEEE/CVF Conference on Computer Vision and Pattern Recognition (CVPR)},
    month     = {June},
    year      = {2026},
    pages     = {9689-9698}
}

@misc{qwen36_27b,
    title = {{Qwen3.6-27B}: Flagship-Level Coding in a 27B Dense Model},
    url = {https://qwen.ai/blog?id=qwen3.6-27b},
    author = {{Qwen Team}},
    month = {April},
    year = {2026}
}

@misc{qwen36_35b_a3b,
    title = {{Qwen3.6-35B-A3B}: Agentic Coding Power, Now Open to All},
    url = {https://qwen.ai/blog?id=qwen3.6-35b-a3b},
    author = {{Qwen Team}},
    month = {April},
    year = {2026}
}

@inproceedings{somers2024kpr,
    author="Somers, Vladimir
    and Alahi, Alexandre
    and Vleeschouwer, Christophe De",
    editor="Leonardis, Ale{\v{s}}
    and Ricci, Elisa
    and Roth, Stefan
    and Russakovsky, Olga
    and Sattler, Torsten
    and Varol, G{\"u}l",
    title="Keypoint Promptable Re-Identification",
    booktitle="Computer Vision -- ECCV 2024",
    year="2025",
    publisher="Springer Nature Switzerland",
    address="Cham",
    pages="216--233",
    isbn="978-3-031-72986-7"
}

@inproceedings{irra,
  author={Jiang, Ding and Ye, Mang},
  booktitle={2023 IEEE/CVF Conference on Computer Vision and Pattern Recognition (CVPR)},
  title={Cross-Modal Implicit Relation Reasoning and Aligning for Text-to-Image Person Retrieval},
  year={2023},
  volume={},
  number={},
  pages={2787-2797},
  doi={10.1109/CVPR52729.2023.00273}
}

@article{kittler1998combining,
  author={Kittler, J. and Hatef, M. and Duin, R.P.W. and Matas, J.},
  journal={IEEE Transactions on Pattern Analysis and Machine Intelligence},
  title={On combining classifiers},
  year={1998},
  volume={20},
  number={3},
  pages={226-239},
  doi={10.1109/34.667881}
}

@article{jain2005score,
    title = {Score normalization in multimodal biometric systems},
    journal = {Pattern Recognition},
    volume = {38},
    number = {12},
    pages = {2270-2285},
    year = {2005},
    issn = {0031-3203},
    doi = {https://doi.org/10.1016/j.patcog.2005.01.012},
    url = {https://www.sciencedirect.com/science/article/pii/S0031320305000592},
    author = {Anil Jain and Karthik Nandakumar and Arun Ross}
}

@inproceedings{icsm,
  author={Pang, Zhiqi and Wang, Junjie and Zhao, Lingling and Wang, Chunyu},
  booktitle={2025 IEEE/CVF Conference on Computer Vision and Pattern Recognition (CVPR)},
  title={Identity-Clothing Similarity Modeling for Unsupervised Clothing Change Person Re-Identification},
  year={2025},
  volume={},
  number={},
  pages={19251-19260},
  doi={10.1109/CVPR52734.2025.01793}
}

@inproceedings{ccfa,
  title={Clothing-change feature augmentation for person re-identification},
  author={Han, Ke and Gong, Shaogang and Huang, Yan and Wang, Liang and Tan, Tieniu},
  booktitle={Proceedings of the IEEE/CVF conference on computer vision and pattern recognition},
  pages={22066--22075},
  year={2023}
}

@inproceedings{chebgr,
  author={Yang, Jinxi and Li, He and Du, Bo and Ye, Mang},
  booktitle={2025 IEEE/CVF Conference on Computer Vision and Pattern Recognition (CVPR)},
  title={Cheb-GR: Rethinking k-nearest neighbor search in Re-ranking for Person Re-identification},
  year={2025},
  volume={},
  number={},
  pages={19261-19270},
  doi={10.1109/CVPR52734.2025.01794}
}

@inproceedings{aslam2001metasearch,
    author = {Aslam, Javed A. and Montague, Mark},
    title = {Models for metasearch},
    year = {2001},
    isbn = {1581133316},
    publisher = {Association for Computing Machinery},
    address = {New York, NY, USA},
    url = {https://doi.org/10.1145/383952.384007},
    doi = {10.1145/383952.384007},
    booktitle = {Proceedings of the 24th Annual International ACM SIGIR Conference on Research and Development in Information Retrieval},
    pages = {276–284},
    numpages = {9},
    location = {New Orleans, Louisiana, USA},
    series = {SIGIR '01}
}

@article{clipcaf,
  title={Clip-driven cloth-agnostic feature learning for cloth-changing person re-identification},
  author={Li, Shuang and Leng, Jiaxu and Li, Guozhang and Gan, Ji and Gao, Xinbo and others},
  journal={arXiv preprint arXiv:2406.09198},
  year={2024}
}

@article{ccpa,
  title={Ccpa: Long-term person re-identification via contrastive clothing and pose augmentation},
  author={Nguyen, Vuong D and Shah, Shishir K},
  journal={arXiv preprint arXiv:2402.14454},
  year={2024}
}

@article{clipscgi,
  title={CLIP-SCGI: synthesized caption-guided inversion for person re-identification},
  author={Han, Qianru and He, Xinwei and Liu, Zhi and Liu, Sannyuya and Zhang, Ying and Xiang, Jinhai},
  journal={arXiv preprint arXiv:2410.09382},
  year={2024}
}

@article{skeldyn,
  title={Clothes-changing person re-identification based on skeleton dynamics},
  author={Joseph, Asaf and Peleg, Shmuel},
  journal={arXiv preprint arXiv:2503.10759},
  year={2025}
}

@article{som2023,
  title={Set-of-mark prompting unleashes extraordinary visual grounding in gpt-4v},
  author={Yang, Jianwei and Zhang, Hao and Li, Feng and Zou, Xueyan and Li, Chunyuan and Gao, Jianfeng},
  journal={arXiv preprint arXiv:2310.11441},
  year={2023}
}

@article{paruchuri2026same,
  title={Same Evidence, Different Answer: Auditing Order Sensitivity in Multimodal Large Language Models},
  author={Paruchuri, Akshay and Koyejo, Sanmi and Adeli, Ehsan},
  journal={arXiv preprint arXiv:2606.26079},
  year={2026}
}

@article{kaplan2020scaling,
  title={Scaling laws for neural language models},
  author={Kaplan, Jared and McCandlish, Sam and Henighan, Tom and Brown, Tom B and Chess, Benjamin and Child, Rewon and Gray, Scott and Radford, Alec and Wu, Jeffrey and Amodei, Dario},
  journal={arXiv preprint arXiv:2001.08361},
  year={2020}
}
}

\newpage

\appendix
\renewcommand{\contentsname}{Technical Appendices and Supplementary Material}
\section{Technical Appendices and Supplementary Material}

\startcontents[sections]
\printcontents[sections]{l}{2}{\setcounter{tocdepth}{3}}
\vspace{2cm}
\hrule
\vspace{1cm}

\subsection{Dataset statistics}
\label{sec:dataset-stats}
\cref{tab:datasets} sizes the four CHIRLA scenarios and reports how many
train\_$k$/test\_$k$ subset pairs each one evaluates, since metrics are averaged over pairs rather than pooled over queries, and the pairs are uneven. Subset $0$ is the designated validation subset used for inference-time configuration selection and is excluded from every reported metric.

\begin{table}[h]
\centering
\caption{\textbf{CHIRLA statistics.} \emph{Subsets} counts the train\_$k$/test\_$k$ pairs
the protocol evaluates, excluding subset \texttt{0}. Sizes are summed over the pairs;
identity and camera counts are unions over them. \emph{Closed} counts the queries whose
identity is not a distractor and appears in the corresponding gallery, and the gap to
\emph{all} is what the closed-set protocol discards.}
\label{tab:datasets}
\setlength{\tabcolsep}{5pt}
\small
\begin{tabular}{lccccccc}
\toprule
\multirow{2}{*}{Scenario} & \multirow{2}{*}{Subsets} & \multicolumn{3}{c}{Query} & \multicolumn{2}{c}{Gallery} & \multirow{2}{*}{Cams.} \\
\cmidrule(lr){3-5} \cmidrule(lr){6-7}
 & & All & Closed & IDs & Imgs. & IDs & \\
\midrule
LT & 6 & 4903 & 3692 & 10 & 368 & 13 & 6 \\
MC-LT & 9 & 2207 & 1134 & 11 & 252 & 19 & 7 \\
MC & 9 & 4454 & 4311 & 19 & 305 & 19 & 7 \\
Reapp. & 9 & 1251 & 1251 & 21 & 258 & 21 & 4 \\
\bottomrule
\end{tabular}
\end{table}

\subsection{FLOPs calculation}
\label{sec:flops-accounting}
We account for compute as
\begin{equation}
\text{FLOPs} \;\approx\; \text{FLOPs}_\text{embed} \;+\; 2 N_\text{act} D,
\label{eq:flops}
\end{equation}
the standard forward-pass count of one multiply-add per parameter per
token~\cite{kaplan2020scaling}, summing the experts' forward passes over queries and gallery with the judge's cost over the $D$ prompt and completion tokens it processes.
$N_\text{act}$ is the judge's active parameter count ($27$B for the dense model used throughout the compute analysis). Per-image forward costs for the experts are measured on CHIRLA crops.

Two terms are omitted. The first is attention's context-dependent cost, given in the same reference as $2\,n_\text{layer}\,n_\text{ctx}\,d_\text{attn}$ per token; for the served configuration ($64$ layers, $d_\text{attn} = 6144$) at our measured token counts this is $2.2\%$ of the parameter term at $P = 6$, rising to $7.9\%$ at $P = 40$. The second is the vision tower that converts crops into image tokens; those tokens are counted once they reach the language model, but the encoder pass producing them is not. Both omissions scale with context and with the number of images, so at a fixed $P$ they inflate every pipeline near-equally and leav e the ratios between them intact, and across $P$ they understate how fast cost grows. Under \cref{eq:flops} cost is linear in $P$, since $D$ is; the growth the omitted attention term would add is what makes the true cost superlinear.

\subsection{Ensemble baselines}
\label{sec:ensembles}
The three aggregation baselines of \cref{tab:main_headline} combine the two experts' full rankings of the gallery, not a shortlist. \emph{Voting} is a Borda count~\cite{aslam2001metasearch}: each expert awards a gallery item $N - \text{rank}$ points and the totals are summed. \emph{Score Fusion} min--max normalises each expert's similarities to $[0, 1]$ and averages them~\cite{jain2005score}. \emph{RRF}~\cite{rrf2009} sums $1/(k + \text{rank})$ across experts with $k = 60$. All three are computed from the raw embeddings rather than the saved top-$K$ rankings, so no truncation is involved.

\subsection{Full results over all pool budgets}
\label{sec:full-results}
\cref{tab:all_budgets} extends our evaluation to $P \in \{6, 10, 16, 20, 30, 40\}$, includes SapiensID + VLM and Union + VLM\qual{all} results, and FLOPs calculations.

\begin{table}[t]
\centering
\caption{CHIRLA Results over $P$. VLM: \textit{Qwen3.6-27B}. In \textbf{bold} best result per column and methods grouped. PFLOPs is the total forward compute to run all four benchmarks once, summed over scenarios: backbone passes over the closed-set queries and gallery, plus $2 N_\text{act} D$ for the $D$ tokens the VLM processes ($N_\text{act}{=}27$B active parameters).}
\label{tab:all_budgets}
\resizebox{\linewidth}{!}{%
\begin{tabular}{lcccccccc|cc|c}
\toprule
\multirow{2}{*}{Method} & \multicolumn{2}{c}{LT} & \multicolumn{2}{c}{MC-LT} & \multicolumn{2}{c}{MC} & \multicolumn{2}{c}{Reapp.} & \multicolumn{2}{c}{Avg.} & \multirow{2}{*}{PFLOPs} \\
\cmidrule(lr){2-3} \cmidrule(lr){4-5} \cmidrule(lr){6-7} \cmidrule(lr){8-9} \cmidrule(lr){10-11}
 & R-1 & mAP & R-1 & mAP & R-1 & mAP & R-1 & mAP & R-1 & mAP & \\
\midrule
DIFFER~\cite{liang2025differ} & 30.89 & 31.20 & 42.56 & \best{48.51} & 52.67 & 53.79 & 54.15 & 55.45 & 45.07 & 47.24 & 1.9 \\
SapiensID~\cite{kim2025sapiensid} & \best{35.50} & \best{35.01} & \best{43.44} & 46.63 & \best{61.28} & \best{57.94} & \best{65.50} & \best{61.22} & \best{51.43} & \best{50.20} & 0.6 \\
\midrule

Voting \qual{Borda}~\cite{aslam2001metasearch} & 38.63 & 36.17 & 45.81 & 50.27 & 60.33 & 59.27 & 62.84 & 61.00 & 51.90 & 51.68 & 2.4 \\
RRF~\cite{rrf2009} & 39.60 & 37.41 & \best{49.49} & \best{52.14} & 61.06 & 60.47 & 63.20 & 62.31 & 53.34 & 53.08 & 2.4 \\
Score Fusion~\cite{jain2005score} & \best{40.95} & \best{38.31} & 48.60 & 51.78 & \best{64.47} & \best{61.75} & \best{65.93} & \best{63.90} & \best{54.99} & \best{53.94} & 2.4 \\
\midrule

SapiensID + VLM \PPs{6} & 42.85 & 37.60 & 53.43 & 51.22 & 71.60 & 62.54 & 72.59 & 64.30 & 60.11 & 53.91 & 954 \\
SapiensID + VLM \PPs{10} & 44.63 & 39.77 & 56.92 & 55.22 & 74.60 & 66.99 & 74.41 & 66.72 & 62.64 & 57.18 & 1207 \\
SapiensID + VLM \PPs{16} & 44.88 & 41.82 & 57.36 & 59.53 & 77.63 & 71.94 & 76.25 & 71.66 & 64.03 & 61.24 & 1592 \\
SapiensID + VLM \PPs{20} & 44.76 & 42.71 & 57.79 & 62.02 & 78.21 & 73.60 & 76.38 & 74.07 & 64.28 & 63.10 & 1861 \\
SapiensID + VLM \PPs{30} & 44.28 & 44.28 & \best{58.79} & \best{63.88} & \best{78.31} & 75.98 & 75.97 & 77.00 & \best{64.34} & 65.29 & 2475 \\
SapiensID + VLM \PPs{40} & \best{44.98} & \best{45.77} & 56.29 & 63.21 & 77.82 & \best{78.96} & \best{77.29} & \best{78.57} & 64.09 & \best{66.63} & 3079 \\
\midrule

Union + VLM \qual{gated} \PPs{6} & 44.97 & 39.59 & 56.35 & 54.43 & 73.25 & 64.87 & 73.72 & 65.28 & 62.07 & 56.04 & 594 \\
Union + VLM \qual{gated} \PPs{10} & 45.95 & 41.02 & \best{59.04} & 56.66 & 74.35 & 66.60 & 75.13 & 66.04 & 63.62 & 57.58 & 728 \\
Union + VLM \qual{gated} \PPs{16} & 46.92 & 42.91 & 56.99 & 57.49 & 76.10 & 69.16 & \best{77.41} & 69.11 & \best{64.35} & 59.67 & 921 \\
Union + VLM \qual{gated} \PPs{20} & \best{47.01} & 43.51 & 56.02 & 57.53 & 75.46 & 69.84 & 76.60 & 69.16 & 63.77 & 60.01 & 1057 \\
Union + VLM \qual{gated} \PPs{30} & 46.30 & 44.09 & 55.73 & 59.21 & 77.30 & 71.88 & 76.40 & 70.15 & 63.93 & 61.33 & 1397 \\
Union + VLM \qual{gated} \PPs{40} & 46.05 & \best{44.75} & 55.47 & \best{59.24} & \best{77.68} & \best{73.14} & 76.53 & \best{71.49} & 63.93 & \best{62.15} & 1694 \\
\midrule

Union + VLM \qual{all} \PPs{6} & 45.76 & 40.43 & \best{58.25} & 57.08 & 73.59 & 66.49 & 74.89 & 67.20 & 63.12 & 57.80 & 905 \\
Union + VLM \qual{all} \PPs{10} & 46.40 & 42.44 & 57.81 & 59.03 & 75.83 & 69.97 & 77.00 & 69.77 & 64.26 & 60.30 & 1110 \\
Union + VLM \qual{all} \PPs{16} & 46.25 & 44.00 & 56.60 & 60.61 & 78.67 & 74.03 & \best{78.62} & 73.30 & 65.03 & 62.99 & 1409 \\
Union + VLM \qual{all} \PPs{20} & \best{46.93} & 45.18 & 57.16 & 61.43 & 78.65 & 75.31 & 77.85 & 73.80 & \best{65.15} & 63.93 & 1622 \\
Union + VLM \qual{all} \PPs{30} & 44.98 & 45.64 & 55.55 & 61.91 & \best{79.15} & 77.45 & 76.58 & 75.23 & 64.07 & 65.06 & 2147 \\
Union + VLM \qual{all} \PPs{40} & 44.69 & \best{45.89} & 54.76 & \best{62.75} & 78.93 & \best{78.20} & 77.00 & \best{77.00} & 63.84 & \best{65.96} & 2594 \\

\bottomrule
\end{tabular}
}
\end{table}

\subsection{Uncertainty estimates}
\label{sec:uncertainty}
The pipeline is deterministic, with only variability involved in the set of queries evaluated.
These are video frames from tracklets of CHIRLA's $22$ identities, and frames and tracklets of one person are strongly correlated. For this reason, we resample identities rather than queries.
Each of $10{,}000$ bootstrap replicates draws identities with replacement, shares that draw across all subsets and scenarios, and recomputes each metric as reported; replicates that leave a subset empty are redrawn, and comparisons are paired on the same draw.
Since a bootstrap over so few clusters can give intervals that are too narrow, we also report an exact sign-flip test over identities and call a difference significant only when it gives $p < 0.05$.
Paired intervals are much narrower than single-method ones, a median half-width of $1.5$ against $5.0$ points on Avg.\ R-1, because per-identity difficulty cancels in the difference.
\cref{tab:bootstrap_paired} lists every comparison made in \cref{sec:experiments}.

\begin{table}[t]
\centering
\caption{\textbf{Paired uncertainty for every comparison in \cref{sec:experiments}.}
$\Delta$ is the first method minus the second, in points; for Binding and Removal rows it is the change from adding or removing the named prompt component.
Values are on the Avg.\ column, except the prompt ablations, which were run on Reapp.\ only.
Intervals are 95\% from the identity-level paired bootstrap.
$p$ is the exact sign-flip test over identities and $k$ the number of identities whose share of $\Delta$ is non-zero, so the smallest attainable $p$ is $2/2^k$.
$p < 0.05$ in bold, uncorrected for multiple comparisons.}
\label{tab:bootstrap_paired}
\resizebox{\linewidth}{!}{%
\begin{tabular}{lrrrrrr}
\toprule
\multirow{2}{*}{Comparison} & \multicolumn{3}{c}{R-1} & \multicolumn{3}{c}{mAP} \\
\cmidrule(lr){2-4} \cmidrule(lr){5-7}
 & $\Delta$ [95\% CI] & $p$ & $k$ & $\Delta$ [95\% CI] & $p$ & $k$ \\
\midrule
\multicolumn{7}{l}{\textit{Against SapiensID}} \\
Union + VLM \qual{gated} \PPs{16} vs.\ SapiensID & $+12.9$\,{\scriptsize$[+9.3, +16.8]$} & \textbf{\textless\,0.001} & 18 & $+9.5$\,{\scriptsize$[+7.1, +11.7]$} & \textbf{\textless\,0.001} & 20 \\
Union + VLM \qual{gated} \PPs{6} vs.\ SapiensID & $+10.6$\,{\scriptsize$[+7.0, +14.4]$} & \textbf{\textless\,0.001} & 19 & $+5.8$\,{\scriptsize$[+3.4, +8.1]$} & \textbf{\textless\,0.001} & 20 \\
Score Fusion vs.\ SapiensID & $+3.6$\,{\scriptsize$[+0.3, +7.0]$} & 0.084 & 19 & $+3.7$\,{\scriptsize$[+1.5, +5.9]$} & \textbf{0.012} & 21 \\
RRF vs.\ SapiensID & $+1.9$\,{\scriptsize$[-1.8, +5.8]$} & 0.426 & 19 & $+2.9$\,{\scriptsize$[+0.3, +5.3]$} & 0.078 & 21 \\
Voting \qual{Borda} vs.\ SapiensID & $+0.5$\,{\scriptsize$[-3.2, +4.3]$} & 0.823 & 19 & $+1.5$\,{\scriptsize$[-1.1, +4.0]$} & 0.362 & 21 \\
DIFFER vs.\ SapiensID & $-6.4$\,{\scriptsize$[-12.6, +0.2]$} & 0.105 & 20 & $-3.0$\,{\scriptsize$[-8.0, +1.7]$} & 0.310 & 21 \\
\midrule
\multicolumn{7}{l}{\textit{Against Score Fusion}} \\
Union + VLM \qual{gated} \PPs{16} vs.\ Score Fusion & $+9.4$\,{\scriptsize$[+7.6, +10.7]$} & \textbf{\textless\,0.001} & 19 & $+5.7$\,{\scriptsize$[+4.4, +6.7]$} & \textbf{\textless\,0.001} & 21 \\
Union + VLM \qual{gated} \PPs{6} vs.\ Score Fusion & $+7.1$\,{\scriptsize$[+5.5, +8.1]$} & \textbf{\textless\,0.001} & 20 & $+2.1$\,{\scriptsize$[+0.9, +3.1]$} & \textbf{0.004} & 21 \\
\midrule
\multicolumn{7}{l}{\textit{Pool, gate, and budget}} \\
Pool, union vs.\ SapiensID only, \qual{all} \PPs{6} & $+3.0$\,{\scriptsize$[+0.5, +5.1]$} & 0.054 & 20 & $+3.9$\,{\scriptsize$[+1.4, +6.0]$} & \textbf{0.006} & 21 \\
Pool, union vs.\ SapiensID only, \qual{all} \PPs{16} & $+1.0$\,{\scriptsize$[-0.6, +2.6]$} & 0.264 & 17 & $+1.8$\,{\scriptsize$[-0.6, +4.0]$} & 0.264 & 20 \\
Pool, union vs.\ SapiensID only, \qual{all} \PPs{40} & $-0.2$\,{\scriptsize$[-1.3, +0.8]$} & 0.621 & 19 & $-0.7$\,{\scriptsize$[-1.7, +0.5]$} & 0.212 & 20 \\
Gate, \qual{gated} vs.\ \qual{all} \PPs{6} & $-1.1$\,{\scriptsize$[-1.8, -0.2]$} & 0.161 & 17 & $-1.8$\,{\scriptsize$[-2.1, -1.4]$} & \textbf{\textless\,0.001} & 21 \\
Gate, \qual{gated} vs.\ \qual{all} \PPs{16} & $-0.7$\,{\scriptsize$[-1.8, +0.6]$} & 0.186 & 18 & $-3.3$\,{\scriptsize$[-4.0, -2.4]$} & \textbf{\textless\,0.001} & 20 \\
Gate, \qual{gated} vs.\ \qual{all} \PPs{40} & $+0.1$\,{\scriptsize$[-1.2, +1.6]$} & 0.905 & 20 & $-3.8$\,{\scriptsize$[-4.6, -2.7]$} & \textbf{\textless\,0.001} & 21 \\
Budget, \qual{gated} \PPs{16} vs.\ \PPs{6} & $+2.3$\,{\scriptsize$[+1.5, +3.1]$} & \textbf{\textless\,0.001} & 19 & $+3.6$\,{\scriptsize$[+3.1, +4.2]$} & \textbf{\textless\,0.001} & 20 \\
Budget, \qual{gated} \PPs{40} vs.\ \PPs{16} & $-0.4$\,{\scriptsize$[-1.8, +0.8]$} & 0.550 & 16 & $+2.5$\,{\scriptsize$[+1.5, +3.4]$} & \textbf{\textless\,0.001} & 20 \\
\midrule
\multicolumn{7}{l}{\textit{Tail rule, pool order, and judge}} \\
Tail, majority vs.\ DIFFER & \multicolumn{3}{c}{identical by construction} & $+1.2$\,{\scriptsize$[+0.6, +1.9]$} & \textbf{0.003} & 20 \\
Tail, majority vs.\ SapiensID & \multicolumn{3}{c}{identical by construction} & $+1.3$\,{\scriptsize$[+0.6, +1.9]$} & \textbf{0.002} & 19 \\
Order, $\rho$ (SapiensID, DIFFER) vs.\ (DIFFER, SapiensID) & $+0.7$\,{\scriptsize$[-0.0, +1.6]$} & 0.121 & 17 & $+0.2$\,{\scriptsize$[-0.2, +0.6]$} & 0.436 & 20 \\
Judge, Qwen3.6-27B vs.\ Qwen3.6-35B-A3B & $+1.4$\,{\scriptsize$[+0.4, +2.5]$} & \textbf{0.013} & 18 & $+1.0$\,{\scriptsize$[+0.5, +1.5]$} & \textbf{0.002} & 20 \\
\midrule
\multicolumn{7}{l}{\textit{Prompt ablations (Reapp.)}} \\
Binding, $+$anchor (images before) & $+4.8$\,{\scriptsize$[+3.9, +5.7]$} & \textbf{\textless\,0.001} & 16 & $+2.2$\,{\scriptsize$[+1.7, +2.6]$} & \textbf{\textless\,0.001} & 20 \\
Binding, $+$anchor (images after) & $+8.5$\,{\scriptsize$[+3.5, +13.3]$} & \textbf{\textless\,0.001} & 17 & $+3.3$\,{\scriptsize$[+1.5, +5.0]$} & \textbf{\textless\,0.001} & 20 \\
Binding, $+$label & $+0.7$\,{\scriptsize$[-1.1, +2.6]$} & 0.633 & 15 & $+1.6$\,{\scriptsize$[+1.0, +2.0]$} & \textbf{\textless\,0.001} & 17 \\
Binding, $+$anchor $+$label (images after) & $+9.3$\,{\scriptsize$[+5.0, +12.7]$} & \textbf{\textless\,0.001} & 16 & $+4.8$\,{\scriptsize$[+2.8, +6.8]$} & \textbf{\textless\,0.001} & 20 \\
Position, after vs.\ before (no anchor) & $-1.5$\,{\scriptsize$[-4.5, +1.6]$} & 0.510 & 17 & $-0.1$\,{\scriptsize$[-1.3, +1.0]$} & 0.877 & 20 \\
Position, after vs.\ before (anchor) & $+2.2$\,{\scriptsize$[-0.6, +4.8]$} & 0.323 & 15 & $+1.0$\,{\scriptsize$[-0.1, +1.9]$} & 0.109 & 18 \\
Removal, attribute list (system) & $-1.2$\,{\scriptsize$[-3.4, +0.2]$} & 0.217 & 13 & $-0.8$\,{\scriptsize$[-1.8, +0.1]$} & 0.222 & 17 \\
Removal, attribute list (user) & $-1.0$\,{\scriptsize$[-3.3, +0.9]$} & 0.522 & 17 & $-0.7$\,{\scriptsize$[-1.7, +0.3]$} & 0.532 & 19 \\
Removal, clothing prohibition and escape hatch & $-0.6$\,{\scriptsize$[-1.6, +0.6]$} & 0.728 & 13 & $-0.5$\,{\scriptsize$[-1.2, +0.2]$} & 0.942 & 15 \\
Removal, escape hatch only & $-1.7$\,{\scriptsize$[-3.5, -0.1]$} & 0.116 & 13 & $-0.8$\,{\scriptsize$[-1.7, +0.0]$} & 0.215 & 18 \\
Removal, step-by-step directive & $-0.2$\,{\scriptsize$[-1.3, +1.4]$} & 0.869 & 11 & $-0.4$\,{\scriptsize$[-1.0, +0.5]$} & 0.743 & 15 \\
\bottomrule
\end{tabular}%
}
\end{table}

\subsection{Output-contract failures}
\label{sec:format-failures}
\cref{tab:format_failures} breaks the reliability trend of \cref{sec:exp:design} down by scenario, pipeline and pool budget.
A response is \emph{corrected} when the first attempt was not a valid permutation of $1..P$ and the correction prompt repaired it, and \emph{unrecovered} when it was still invalid after both attempts, in which case the output is discarded and the ranking of $m^\star$ is kept.
From $P = 16$ the rate grows steeply in every row, the one exception being \mbox{MC-LT} under \emph{Union + VLM \qual{gated}}, which falls at $P = 40$.
Below $P = 16$ every rate is under $1\%$ and the small non-monotonicities there are within noise.
Unrecovered failures stay at most $2.6\%$ throughout, so the correction step absorbs nearly all of the growth, at the price of a second and sometimes third call on up to a third of queries.

\begin{table}[t]
\centering
\caption{\textbf{Output-contract failures as a function of $P$.}
Each cell is \emph{corrected / unrecovered}, as a percentage of the queries sent to the judge.
Judge: \textit{Qwen3.6-27B}. See \cref{sec:format-failures}.}
\label{tab:format_failures}
\begin{tabular}{lcccccc}
\toprule
Scenario & $P{=}6$ & $P{=}10$ & $P{=}16$ & $P{=}20$ & $P{=}30$ & $P{=}40$ \\
\midrule
\textit{SapiensID + VLM} &  &  &  &  &  &  \\
\quad LT & 0.2 / 0.1 & 0.2 / 0.1 & 5.3 / 1.5 & 8.9 / 1.6 & 20.9 / 1.7 & 32.0 / 2.6 \\
\quad MC-LT & 0.2 / 0.2 & 0.1 / 0.0 & 3.4 / 0.1 & 4.3 / 0.0 & 7.1 / 0.1 & 9.6 / 0.4 \\
\quad MC & 0.1 / 0.1 & 0.1 / 0.1 & 4.6 / 0.0 & 7.2 / 0.0 & 7.5 / 0.4 & 22.7 / 1.4 \\
\quad Reapp. & 0.1 / 0.4 & 0.4 / 0.0 & 3.7 / 0.0 & 6.8 / 0.0 & 16.2 / 0.3 & 18.9 / 0.5 \\
\midrule
\textit{Union + VLM \qual{gated}} &  &  &  &  &  &  \\
\quad LT & 0.1 / 0.1 & 0.5 / 0.6 & 2.5 / 0.1 & 5.2 / 0.2 & 14.3 / 0.2 & 22.2 / 0.5 \\
\quad MC-LT & 0.7 / 0.3 & 0.3 / 0.1 & 1.8 / 0.5 & 3.3 / 0.3 & 6.0 / 0.4 & 4.7 / 0.1 \\
\quad MC & 0.2 / 0.0 & 0.6 / 0.4 & 1.9 / 0.2 & 4.5 / 0.3 & 12.0 / 0.2 & 16.3 / 0.7 \\
\quad Reapp. & 0.1 / 0.1 & 0.3 / 0.9 & 2.1 / 0.3 & 5.4 / 0.0 & 8.8 / 0.4 & 14.2 / 0.3 \\
\bottomrule
\end{tabular}
\end{table}

\subsection{Prompt templates}
\label{supl:prompts}

\begin{figure}[t]
  \centering
  \begin{subfigure}[b]{0.11\linewidth}
    \centering
    \includegraphics[height=3.6cm]{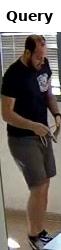}
    \caption{Query}
    \label{fig:prompt_example:query}
  \end{subfigure}
  \hspace{0.04\linewidth}
  \begin{subfigure}[b]{0.47\linewidth}
    \centering
    \includegraphics[height=3.5cm]{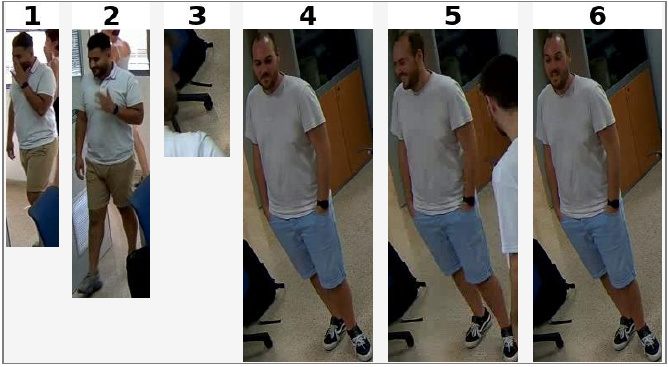}
    \caption{Candidates $1..P$}
    \label{fig:prompt_example:cands}
  \end{subfigure}
  \caption{\textbf{A query and its $P = 6$ pool.} The images are sent as $P + 1$ separate entries after the instructions, in the order $\rho$.
Each crop carries its role in a white banner burned into the pixels, \texttt{Query} or the candidate number, which is the \emph{Label} intervention of \cref{tab:vlm_prompt_ablation}. Without it the role survives only in the
  surrounding text, and the judge has to infer which crop is which from position alone.}
  \label{fig:prompt_example}
\end{figure}

The reranker is driven by three templates. Placeholders in braces are filled at call time: \texttt{\{N\}} is the pool size $P$, and the sentence about banners, shown filled in below, is present only when burned-in labels are used. The example line is a fixed rotation of $1..P$ by two positions, identical for every query, so that it demonstrates the output format without suggesting an ordering. The anchor sentence ablated in \cref{tab:vlm_prompt_ablation} is the
second sentence of the user message, \emph{``The Query is the FIRST image; the last image is Candidate \{N\}, not the Query.''} The third sentence, which announces the burned-in banners, is the separate \emph{Label} intervention. The banner is a $32$ px white strip prepended above the crop, with the label centred in bold (see~\cref{fig:prompt_example}).

\begin{tcolorbox}[promptbox, title=Prompt Template: System Message]
You are an expert in person re-identification across long-term, person-occluded,
cloth-changing, multi-camera surveillance scenarios. The user will show you a Query person
crop and several numbered Candidate crops retrieved by a Re-ID system.

Analyze each person using features that remain stable across clothing changes: body shape,
build, skin tone, gender, and apparent age. Do not rely on clothing color or type as primary
evidence. Nevertheless, if you are unsure, then you can pay attention to the fashion style as
help.

Find similarities among the candidates and group them by high probability of being the same
person.

Reason step by step for each candidate before giving your final answer.
\end{tcolorbox}

\begin{tcolorbox}[promptbox, title=Prompt Template: User Message]
Below you will see a Query person followed by the numbered Candidates retrieved by a
cloth-changing Re-ID system. The Query is the FIRST image; the last image is Candidate
\{N\}, not the Query. Each image has its label printed in a white banner at the top:
``Query'' for the Query person, and its number (1--\{N\}) for each candidate.

Reason step by step about each candidate's similarity to the Query, focusing on body shape,
build, skin tone, gender, and age --- features that remain stable across clothing changes.

Then output a final line with ALL candidates re-ordered from most to least likely to be the
same person as the Query, as a comma-separated list of numbers.

Example for \{N\} candidates: RANKING: 3, 4, \dots, \{N\}, 1, 2
\end{tcolorbox}

Appended to the same conversation when the response does not end in a valid permutation of
$1..P$, for at most two attempts:
\begin{tcolorbox}[promptbox, title=Prompt Template: Correction Message]
Your previous response could not be parsed. Reason: \{REASON\}
Please end your answer with a line in exactly this format:
RANKING: followed by all \{N\} candidate numbers as a comma-separated permutation of
1..\{N\}.
Example for \{N\} candidates: RANKING: 3, 4, \dots, \{N\}, 1, 2
\end{tcolorbox}

The prompt variants ablated in \cref{tab:vlm_prompt_removals} are obtained by removing individual components of these templates: the stable-attribute enumeration in either message, the clothing prohibition and its ``if unsure'' escape hatch, and the step-by-step reasoning directive. The anchor sentence and the banner-label note are ablated separately, in \cref{tab:vlm_prompt_ablation}.

\begin{table}[t]
  \centering
  \caption{Semantic removals from the full prompt, CHIRLA (Reappearance), $P = 6$. Each row removes one component from the full prompt, the last row of \cref{tab:vlm_prompt_ablation}, which is repeated here for reference. $\Delta$ is measured against it. No removal is significant under the sign-flip test (\cref{tab:bootstrap_paired}).}
  \label{tab:vlm_prompt_removals}
  \begin{tabular}{lrrrr}
    \toprule
    Removed component & R-1 & $\Delta$ & mAP & $\Delta$ \\
    \midrule
    None, full prompt & 73.72 & & 65.28 & \\
    \midrule
    Attribute list, system message        & 72.47 & $-1.25$ & 64.46 & $-0.82$ \\
    Attribute list, user message          & 72.74 & $-0.98$ & 64.57 & $-0.71$ \\
    Clothing prohibition and escape hatch & 73.12 & $-0.60$ & 64.82 & $-0.46$ \\
    Escape hatch only                     & 72.05 & $-1.68$ & 64.44 & $-0.84$ \\
    Step-by-step directive                & 73.56 & $-0.16$ & 64.90 & $-0.38$ \\
    \bottomrule
  \end{tabular}
\end{table}

\subsection{Tail-rule ablation}
\label{sec:tail-ablation}
\cref{tab:ablation_ranks_sid} extends the tail-rule ablation of \cref{sec:ablations} with every way of counting model influence.
The first two rows fix the tail to one model.
The rest use the per-query majority vote and differ only in who is credited for a candidate \emph{both} models retrieved, since the merge keeps a single copy and by position it is always attributed to the first-listed model.
``credit own-rank'' credits the model that ranked the candidate higher, and ``credit all'' is the rule \cref{eq:contrib} states, where every retriever counts, which is the only setting in which $c_m$ can tie; ``tie'' names the model that wins such a tie.
We report mAP only, since the head comes from the judge and the tail starts below it, so no tail rule can move R-1.
Every majority variant beats both fixed tails in all four scenarios, and the six differ from each other by at most $0.3$ mAP.

\begin{table}[t]
\centering
\caption{\textbf{Tail rule ablation.} mAP (\%) at $P = 6$ with head size $k_H = k = 3$, per-subset averages, SapiensID as the agreement source. Judge: \textit{Qwen3.6-27B}. See \cref{sec:tail-ablation} for the rules.
\label{tab:ablation_ranks_sid}}
\begin{tabular}{lcccc}
\toprule
Tail rule & LT & MC-LT & MC & Reapp. \\
\midrule
Fixed tail: DIFFER & 38.32 & 54.23 & 63.09 & 63.61 \\
Fixed tail: SapiensID & 38.51 & 52.95 & 63.05 & 64.39 \\
\addlinespace
\textit{Majority tail} & & & & \\
\quad credit DIFFER & 39.49 & \best{54.64} & 64.77 & 64.99 \\
\quad credit SapiensID & \best{39.59} & 54.35 & 64.69 & 65.27 \\
\quad credit own-rank (tie DIFFER) & 39.52 & 54.62 & 64.81 & 65.11 \\
\quad credit own-rank (tie SapiensID) & 39.58 & 54.47 & 64.79 & 65.18 \\
\quad credit all (tie DIFFER) & 39.56 & 54.61 & 64.78 & 65.04 \\
\quad credit all (tie SapiensID) & \best{39.59} & 54.43 & \best{64.87} & \best{65.28} \\
\bottomrule
\end{tabular}
\end{table}

\clearpage

\end{document}